\documentclass{article}

\usepackage[preprint]{neurips_2026}

\usepackage[utf8]{inputenc} 
\usepackage[T1]{fontenc}    
\usepackage{hyperref}       
\usepackage{url}            
\usepackage{booktabs}       
\usepackage{amsfonts}       
\usepackage{nicefrac}       
\usepackage{microtype}      
\usepackage{xcolor}         
\usepackage{pifont}
\usepackage{multirow}
\usepackage{graphicx}
\usepackage{listings}
\usepackage{subcaption}
\usepackage{wrapfig}
\usepackage{xspace}
\usepackage{makecell}

\usepackage{array}
\usepackage{tabularx}

\title{WebArena-Pro: A Heterogeneous, Multimodal, Reproducible Benchmark for Web Agents}

\author{%
  \textbf{%
    Imene Kerboua$^{1\,2\,*\,\dagger}$%
    \begingroup
      \thanks{Equal contribution.
        \enspace $\dagger$ Project leads.
        \enspace $\ddagger$ Core contributors.
        \enspace $\S$ Corresponding authors:
        Xing Han Lù \texttt{<xing.han.lu@mail.mcgill.ca>} and Siva Reddy
        \texttt{<siva.reddy@mila.quebec>}.}%
    \endgroup,
    ~Fatemeh Pesaran Zadeh$^{3\,*\,\dagger}$,
    ~Xing Han Lù$^{4\,5\,\dagger\,\S}$,%
  }\\[2pt]
  \textbf{%
    Weijian Qi$^{6\,\ddagger}$,
    ~Alexander Miller$^{4\,\ddagger}$,
    ~Junyi Song$^{4\,\ddagger}$,
    Yunjia Tian$^{4\,\ddagger}$,
    ~Dongjin Kang$^{3\,\ddagger}$,
  }\\[2pt]
  \textbf{%
    ~Seyeon Choi$^{3\,\ddagger}$,
    Marzia Nouri$^{4\,5}$,
    ~Ewen Gueguen$^{4}$,
    ~Matteo Boglioni$^{5}$,
    Fengyuan Liu$^{4\,5}$,
  }\\[2pt]
  \textbf{%
    ~Zeyi Liao$^{6}$,
    ~Mengqi Yuan$^{7}$,
    Yue Li$^{4}$,
    ~Alexandre Lacoste$^{8}$,
    ~Alexandre Drouin$^{5\,8\,9}$,%
  }\\[2pt]
  \textbf{%
    Spandana Gella$^{4\,5\,8}$,
    ~Huan Sun$^{6}$,
    ~Gunhee Kim$^{3}$,
    ~Siva Reddy$^{4\,5\,8\,10\,\S}$%
  }\\[8pt]
  {\normalfont\small
    $^{1}$INSA Lyon, LIRIS
    \quad $^{2}$Esker
    \quad $^{3}$Seoul National University
  }\\
  {\normalfont\small
    $^{4}$McGill University
    \quad $^{5}$Mila -- Quebec AI Institute
  }\\
  {\normalfont\small
    $^{6}$The Ohio State University
    \quad $^{7}$University of Hong Kong
  }\\
  {\normalfont\small
    $^{8}$ServiceNow AI Research
    \quad $^{9}$Universit\'e Laval
    \quad $^{10}$Canada CIFAR AI Chair
  }%
}

\usepackage{multirow}
\usepackage{pifont}
\newcommand{\cmark}{\textcolor{green!60!black}{\ding{51}}}
\newcommand{\xmark}{\textcolor{red!70!black}{\ding{55}}}
\usepackage{wasysym}

\begin{document}

\maketitle

    \begin{abstract}
    Web agents powered by large language and vision-language models are increasingly applied to realistic browser work that spans heterogeneous applications, multimodal content, and stateful workflows. However, existing reproducible web-agent benchmarks cover only a small number of web applications drawn from a few software categories, and restrict modality to text and vision. Live benchmarks broaden site coverage but sacrifice reproducibility, since pages and data drift between runs. Moreover, existing benchmarks do not meaningfully evaluate whether agents can understand and use audio and video content embedded within web tasks. To address these gaps, we introduce WebArena-Pro, a benchmark comprising 300 tasks across 20 self-hosted web applications in six domain categories, spanning distinct interface conventions, workflows, and data models. Across the evaluated agents, the best performance is achieved by both Codex with GPT 6.1 Sol and Claude Code with Claude Opus 5.5 at 62.3\%, followed by  OpenCode with Kimi 3.1 Pro at 60.7\%. The strongest GenericAgent configurations, Gemini 3.1 Pro omni and Gemma4 31B, both achieve 30.3\%. Among reproducible, human-curated web agent benchmarks, WebArena-Pro provides the broadest application coverage and the most comprehensive multimodal support to date. The benchmark treats audio and video as core observations alongside text and vision, with dedicated actions for extracting information from each. WebArena-Pro runs each task in isolation and supports reproducible, parallel evaluation. Tasks are authored through a dedicated annotator interface, filtered by LLM-assisted triage, and finally validated by humans before release. Project page: \url{https://webarena-pro.github.io}.
\end{abstract}

\section{Introduction}
\label{sec:introduction}

\begin{figure*}[t!]
    \centering
    \includegraphics[width=\textwidth]{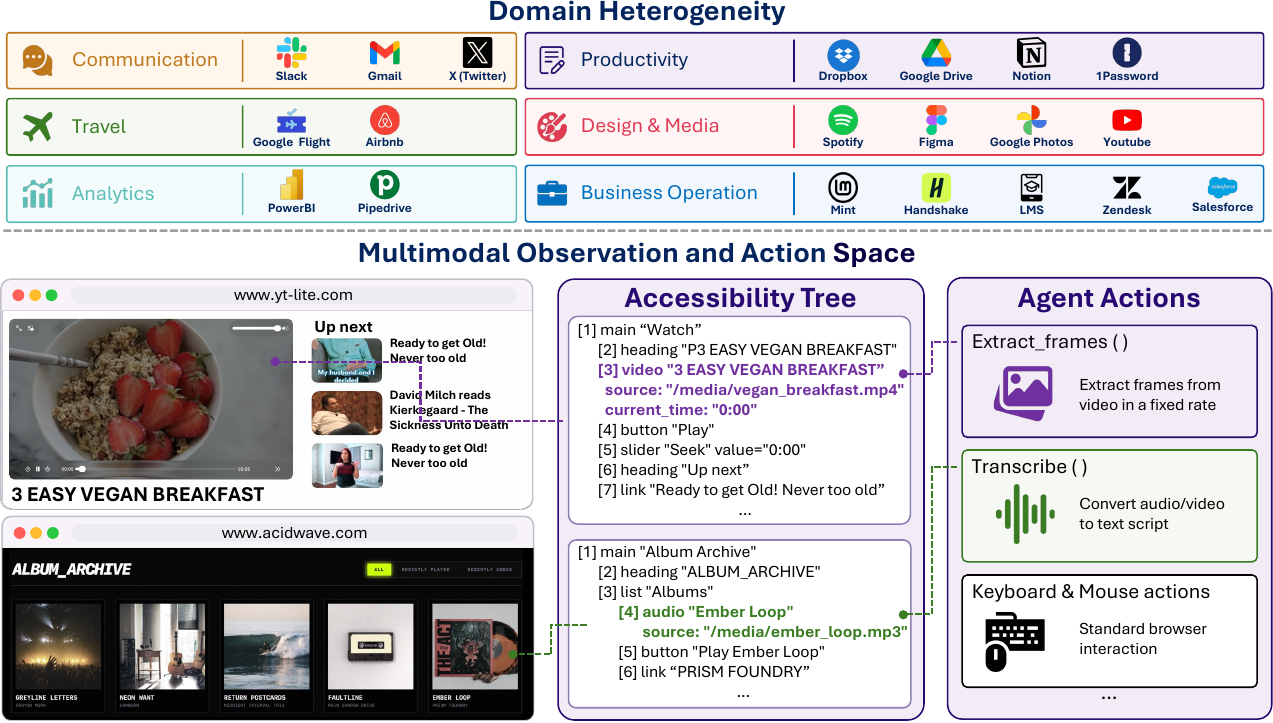}
    \caption{
        \textbf{Overview of WebArena-Pro.}
        \textbf{(A) Domain heterogeneity.} WebArena-Pro spans 20 self-hosted web applications across six domain categories (productivity, communication, analytics, business operations, design and media, and travel), each modeled on a widely used service such as Notion, Slack, Spotify, Figma, or Airbnb. The full set of applications and their reference services are listed in Table~\ref{tab:websites}.
        \textbf{(B) Multimodal observation and action space.} WebArena-Pro extends the agent's observation space beyond text and vision to include audio and video. Embedded media sources are exposed to the agent through the accessibility tree, and agents invoke \texttt{transcribe()} and \texttt{extract\_frames()} actions to convert temporal media into a form usable during browser workflows. If the agents handle native audio and video processing, they invoke \texttt{listen()} and \texttt{watch()} actions.
    }
    \label{fig:overview}
\end{figure*}

Web agents driven by large language and vision-language models have moved quickly from synthetic clicking tasks toward broader browser work, including drafting and triaging messages, updating records in business applications, extracting insight from dashboards, editing media, and coordinating across heterogeneous tools \citep{abuelsaadAgentEAutonomousWeb2024, dengMind2WebGeneralistAgent2023, gurRealWorldWebAgentPlanning2023,heWebVoyagerBuildingEndtoEnd2024}. A growing body of benchmarks has supported this progress, offering \textit{self-hosted} web environments \citep{zhouWebArenaRealisticWeb2023, hattami2025webarena}, visually grounded tasks \citep{kohVisualWebArenaEvaluatingMultimodal2024, miyai2025webchorearenaevaluatingwebbrowsing}, workplace-style scenarios \citep{drouinWorkArenaHowCapable2024, xu2025theagentcompanybenchmarkingllmagents}, computer-use settings beyond the browser \citep{osworld2024neurips}, and increasingly large task pools \citep{aggarwal2026gymanythingturnsoftwareagent}. As models improve and the scope of attempted tasks widens, the design of the evaluation itself becomes a central question.

Despite this progress, current self-hosted web-agent benchmarks have limitations relative to the conditions needed for realistic and diverse measurement, as summarized in Table~\ref{tab:compare}. First, prior self-hosted benchmarks cover only a small number of web applications drawn from a few software categories, which limits the heterogeneity of interfaces, workflows, and data models that an agent encounters during evaluation. Second, modality coverage is limited to text and vision, even though many real browser workflows involve temporal media such as audio and video embedded in web pages.
A parallel line of live benchmarks operates on real production websites \citep{dengMind2WebGeneralistAgent2023, xue2025illusion, heWebVoyagerBuildingEndtoEnd2024}, broadening site coverage beyond what self-hosted benchmarks offer. However, they sacrifice reproducibility, since pages, content, layouts, and policies drift between runs, making consistent evaluation across rollouts difficult.

To this end, we introduce \textit{WebArena-Pro}, a benchmark for web agents that addresses these limitations. First, addressing the application coverage gap, WebArena-Pro spans 20 self-hosted web applications across six domain categories (productivity, communication, analytics, business operations, design and media, and travel), with applications modeled on widely used services such as Notion, Slack, Spotify, Figma, and Airbnb (Table~\ref{tab:websites} lists all 20). Second, addressing the modality gap, WebArena-Pro evaluates audio and video as core observations alongside text and vision, exposing media sources through the accessibility tree and providing \texttt{transcribe()} and \texttt{extract\_frames()} actions so that agents can incorporate temporal media directly into browser workflows. Third, addressing the reproducibility limitation of live benchmarks, WebArena-Pro runs each task in isolation against per-task seeded container state, supporting reproducible parallel evaluation that avoids drift between runs.

WebArena-Pro contains 300 tasks categorized by site coverage (204 single-site, 96 cross-site) and modality (221 text/vision, 79 audio/video), as detailed in Table~\ref{tab:taskcats}. Unlike many prior benchmarks that generate many task instances from a small set of templates, each task in WebArena-Pro has its own dedicated template, which yields a more diverse evaluation (Table~\ref{tab:compare}). Tasks are authored through a dedicated annotator interface and pass through a two-stage filter, in which LLM-assisted triage flags candidates that appear infeasible or trivial and human reviewers make the final acceptance decision. This pipeline keeps task quality grounded in human judgment while using LLM assistance to scale the authoring process.
Our contributions are as follows:

\begin{itemize}\itemsep=0pt
\item \textbf{Heterogeneous web coverage.} We construct a benchmark of 20 self-hosted web applications across six domain categories with distinct interface conventions, workflows, and data models.
\item \textbf{Audio and video as core web observations.} We extend the web-agent action and observation space to temporal media, exposing audio and video sources through the accessibility tree and providing \texttt{transcribe()} and \texttt{extract\_frames()} actions so that agents can use embedded media directly within browser workflows. Omni agents are provided with \texttt{listen()} and \texttt{watch()} actions.
\item \textbf{Reproducible, parallel evaluation.} We run each task in isolation against per-task seeded container state, supporting reproducible parallel rollouts and avoiding drift between runs.
\item \textbf{Human-validated benchmark construction.} We release a task creation pipeline that combines an annotator interface, LLM-assisted infeasibility and triviality triage, and final human validation.
\end{itemize}

\section{Related Work}

\begin{table*}[t]
    \centering
    \caption{\small Comparing WebArena-Pro to existing reproducible benchmarks for web and computer-use agents. ``\textit{Isolation}'' refers to per-task environment reset enabling contamination-free parallel execution. ``\textit{Human-curated}'' indicates that tasks are authored or verified by humans rather than auto-generated. A half-circle (\protect\LEFTcircle) indicates partial support. \textbf{$^\S$}WorkArena Suite combines WorkArena (33 templates instantiated to 19{,}912 instances) and WorkArena++ (682 compositional tasks, each sampling from thousands of valid configurations). \textbf{$^\dagger$}OSWorld targets desktop applications rather than websites, with ``\# Envs'' showing its 9 application domains. \textbf{$^\ddagger$}Gym-Anything's environments and tasks are LLM-generated at scale rather than human-curated. }
    \label{tab:compare}
    \setlength{\tabcolsep}{3pt}
    \resizebox{\linewidth}{!}{
    \begin{tabular}{lc cc ccc cc}
        \toprule
        & & \multicolumn{2}{c}{\textbf{Tasks}} & \multicolumn{3}{c}{\textbf{Modalities}} & \multicolumn{2}{c}{\textbf{Eval \& Execution}} \\
        \cmidrule(lr){3-4} \cmidrule(lr){5-7} \cmidrule(lr){8-9}
        \textbf{Benchmark} & \textbf{\# Envs} & \textbf{Templates} & \textbf{Instances} & \textbf{Vision} & \textbf{Audio} & \textbf{Video} & \textbf{Isolation} & \textbf{Human-curated} \\
        \midrule
        WebArena \citep{zhouWebArenaRealisticWeb2023} & 6 & 241 & 812 & \xmark & \xmark & \xmark & \xmark & \cmark \\
        VisualWebArena \citep{kohVisualWebArenaEvaluatingMultimodal2024} & 3 & 233 & 910 & \cmark & \xmark & \xmark & \xmark & \cmark \\
        WebChoreArena \citep{miyai2025webchorearenaevaluatingwebbrowsing} & 4 & 117 & 532 & \cmark & \xmark & \xmark & \xmark & \cmark \\
        WorkArena Suite$^\S$ \citep{boisvertWorkArenaCompositionalPlanning2024,drouinWorkArenaHowCapable2024} & 1 & 715 & 19{,}912 & \cmark & \xmark & \xmark & \LEFTcircle & \cmark \\
        TheAgentCompany \citep{xu2025theagentcompanybenchmarkingllmagents} & 4 & 175 & 175 & \cmark & \xmark & \xmark & \cmark & \cmark \\
        OSWorld$^\dagger$ \citep{osworld2024neurips} & 9 & 369 & 369 & \cmark & \xmark & \xmark & \cmark & \cmark \\
        Gym-Anything$^\ddagger$ \citep{aggarwal2026gymanythingturnsoftwareagent} & 200 & 10{,}000+ & 10{,}000+ & \cmark & \xmark & \xmark & \cmark & \xmark \\
        \midrule
        \textbf{WebArena-Pro} (ours) & 20 & 300 & 300 & \cmark & \cmark & \cmark & \cmark & \cmark \\
        \bottomrule
    \end{tabular}
    }
\end{table*}

\paragraph{Self-Hosted Web Benchmarks.}

WebArena is the closest predecessor to WebArena-Pro. It introduced a self-hosted benchmark for realistic web navigation over six web applications, with agents completing natural-language tasks by operating a browser \citep{zhouWebArenaRealisticWeb2023}. VisualWebArena \citep{kohVisualWebArenaEvaluatingMultimodal2024} extended the setting to visually grounded tasks where success depends on understanding screenshots rather than only DOM text, and WebChoreArena \citep{miyai2025webchorearenaevaluatingwebbrowsing} stressed long-horizon, tedious browsing patterns with self-hosted websites as well. While these benchmarks provide valuable foundations for realistic and reproducible web-agent evaluation, WebArena-Pro differs along three axes visible in Table~\ref{tab:compare}. First, it covers 20 applications across six domain categories, the broadest application coverage among self-hosted web benchmarks. Second, it enforces per-task isolation by explicitly resetting application state between tasks. In contrast, these benchmarks retain state across tasks, so backend changes an agent makes while solving one task persist into the next and can contaminate later evaluations. Third, it authors each task as its own dedicated template. In contrast, these benchmarks instantiate many task instances from a smaller template set (e.g., 241 templates for 812 instances in WebArena), which may lead to a less diverse evaluation.

\paragraph{Enterprise Benchmarks.}
A parallel line of work pushes toward workplace realism. WorkArena evaluates agents on common knowledge-work tasks in ServiceNow \citep{drouinWorkArenaHowCapable2024}, and WorkArena++ extends this direction with compositional planning tasks and automatically generated ground-truth trajectories \citep{boisvertWorkArenaCompositionalPlanning2024}. TheAgentCompany simulates a small company across a handful of internal services and introduces non-binary partial scoring \citep{xu2025theagentcompanybenchmarkingllmagents}. These benchmarks emphasize enterprise workflows, long-horizon planning, and task composition. WebArena-Pro is complementary; rather than centering on one enterprise platform, it spans productivity, communication, analytics, business operations, design and media, and travel applications, while still retaining stateful authenticated workflows.

\paragraph{Live and Computer-Use Benchmarks.}
Other benchmarks broaden the evaluation setting in a different direction. Mind2Web studies generalization to real websites using collected human demonstrations \citep{dengMind2WebGeneralistAgent2023}, and OnlineMind2Web \citep{xue2025illusion} considers a live variant where the target websites may change over time. Live benchmarks improve ecological validity but make repeatability difficult, since pages, data, policies, and layouts can drift between evaluations. WebArena-Pro instead uses self-hosted applications and seeded data, which makes failures easier to reproduce and enables backend-state evaluation. Beyond the browser, computer-use benchmarks such as OSWorld \citep{osworld2024neurips} broaden the target from browsers to full desktop environments, and Gym-Anything \citep{aggarwal2026gymanythingturnsoftwareagent} pursues a different scaling axis by using LLMs to configure environments and auto-generate evaluators. WebArena-Pro instead focuses specifically on web software, where browser instrumentation and application-level reset can be made reliable enough for controlled experiments, and keeps task authoring human-validated.

\paragraph{Multimodal Web Benchmarks.}

VisualWebArena showed that many web tasks require visual grounding and that text-only agents miss information that is obvious to humans \citep{kohVisualWebArenaEvaluatingMultimodal2024}. 
WebArena-Pro follows this motivation and extends multimodal support beyond static visual content to include audio and video. To enable this, we add new actions to the action space of the agent, such as \texttt{transcribe()} and \texttt{extract\_frames()}. The agent invokes \texttt{transcribe()} to convert an audio source to text and \texttt{extract\_frames()} to sample representative frames from a video source. The resulting transcripts are merged into the accessibility-tree observation as synthetic nodes, while extracted frames are returned as image content alongside the AXTree, so the agent perceives temporal media within the same observation structure it already uses for the rest of the page (Figure~\ref{fig:overview}).
Finally, VideoWebArena~\citep{jang2025videowebarena} keeps the original WebArena websites and adds external tutorial videos that agents must draw facts or skills from. Unlike VideoWebArena, WebArena-Pro builds its audio and video content into the 20 new self-hosted applications, so multimodal understanding is part of the environment itself.

\section{WebArena-Pro}

\begin{table*}[t]
    \centering
    \caption{The 20 self-hosted web applications in WebArena-Pro, grouped by domain.
    Each application is a self-hosted instance whose interface and workflow patterns
    are modeled on a widely-used reference application, while data and exact feature
    scope are controlled by the benchmark.}
    \label{tab:websites}
    \setlength{\tabcolsep}{6pt}
    \renewcommand{\arraystretch}{1.1}
    \resizebox{\linewidth}{!}{
    \begin{tabular}{llll}
        \toprule
        \textbf{Domain} & \textbf{Site} & \textbf{Reference application} & \textbf{Description} \\
        \midrule
        Productivity      & MyDrive          & Dropbox            & Personal cloud file storage and sharing \\
                          & NextCloud        & Google Drive       & Collaborative file storage with calendars, contacts, and chat \\
                          & Notion           & Notion             & Notes and knowledge-management workspace \\
                          & Password-Manager & 1Password          & Password and credential vault \\
        \cmidrule(lr){1-1}\cmidrule(lr){2-4}
        Communication     & Mattermost       & Slack              & Team messaging with channels and direct messages \\
                          & Thunderbird      & Gmail              & Email client with folders and contacts \\
                          & X-Lite           & X (Twitter)        & Microblogging and social posting \\
        \cmidrule(lr){1-1}\cmidrule(lr){2-4}
        Analytics         & Metabase         & Power BI           & Business-intelligence dashboards and SQL queries \\
                          & PipeCRM          & Pipedrive          & Sales pipeline and contact management \\
        \cmidrule(lr){1-1}\cmidrule(lr){2-4}
        Business ops.     & ERP-Next         & Salesforce         & Customer and supplier chain management \\
                          & Firefly          & Mint               & Personal finance and budget management \\
                          & JobPortal        & Handshake          & Job postings and application tracking \\
                          & Minerva-Lite     & University LMS     & Courses, assignments, and grades \\
                          & Peppermint       & Zendesk            & Customer-support ticketing \\
        \cmidrule(lr){1-1}\cmidrule(lr){2-4}
        Design \& media   & AcidWave         & Spotify            & Music streaming with lyrics and playlists \\
                          & Immich           & Google Photos      & Photo library with albums and search \\
                          & Penpot           & Figma              & Web-based design and prototyping \\
                          & YT-Lite          & YouTube            & Video hosting and playback \\
        \cmidrule(lr){1-1}\cmidrule(lr){2-4}
        Travel & Flight-Search   & Google Flights     & Flight search and itinerary planning \\
               & TravelMap       & Airbnb             & Travel and lodging booking \\
        \bottomrule
    \end{tabular}
    }
\end{table*}

\subsection{Web Environments}
\label{sec:webapps}

WebArena-Pro consists of 20 self-hosted web applications grouped into six domain categories, listed in Table~\ref{tab:websites}. Each application runs as a Docker container with seeded data and fixed login credentials, and is reachable at a fixed local URL during evaluation.  Before each task, the container state is reset to its seeded baseline, so changes made by an agent during one task do not persist into the next. Local hosting also fixes the application version and the database contents across runs, which removes the drift in pages, data, or policies.

The six domains are productivity, communication, analytics, business operations, design and media, and travel. Within each domain the applications share a broad task family, such as file management in productivity or dashboard analysis in analytics, but differ in interface conventions, authentication flows, and backend state schemas. NextCloud and MyDrive both store files, for example, but they expose distinct REST APIs, distinct folder and permission models, and distinct Document Object Model (DOM) structures, so an agent that has learned to operate one is not guaranteed to transfer to the other. This grouping is what allows WebArena-Pro to test transfer at the level of interface and backend conventions, rather than memorization of a single product.

Each application is modeled on a familiar reference application, listed in the third column of Table~\ref{tab:websites}, which names the workflow pattern reproduced rather than implying data or feature parity. 

\subsection{Tasks and Action Space}
\label{sec:tasks}

\begin{table*}[t]
    \centering
    \small
    \caption{Qualitative examples of WebArena-Pro tasks, organized by site coverage (single-site or cross-site) and modality (text/vision or audio/video).}
    \label{tab:taskcats}
    \setlength{\tabcolsep}{6pt}
    \renewcommand{\arraystretch}{1.2}
    \begin{tabularx}{\linewidth}{llp{4.5em}X}
        \toprule
        \textbf{Coverage} & \textbf{Modality} & \textbf{Sites} & \textbf{Example intent} \\
        \midrule
        Single-site & Text/Vision & Flight-Search & {Set the displayed currency to USD. Find the fastest one-way economy flight from New York (JFK) to Los Angeles (LAX) on 2026-06-15. If multiple flights are tied for the shortest total travel time, choose the first tied flight shown after sorting by Fastest.} \\
        \cmidrule(lr){2-4}
                    & Audio/Video & YT-Lite &{In the video "Tips for winter biking", does the creator record himself from a first person point of view?} \\
        \midrule
        Cross-site  & Text/Vision & Notion, Peppermint & {Create a Notion page summarizing the VPN drops ticket and naming its assigned owner.} \\
        \cmidrule(lr){2-4}
                    & Audio/Video & Mattermost,  Acidwave & {Send Mike a direct message recommending a rock song suitable for someone going through a breakup.} \\
        \bottomrule
    \end{tabularx}
\end{table*}

A task in WebArena-Pro is specified by a natural-language intent, one or more starting sites, an authenticated browser state, and an evaluation specification that determines success and partial credit (Section~\ref{sec:evaluation}). WebArena-Pro contains 300 tasks, with each task authored as its own dedicated template (Table~\ref{tab:compare}).

We categorize these tasks along two independent dimensions, namely site coverage and modality, with qualitative examples in Table~\ref{tab:taskcats}. Along site coverage, 204 tasks are \textbf{single-site} and require the agent to complete a workflow within one application, while 96 are \textbf{cross-site} and require the agent to coordinate between two or more applications, probing transfer across the interface and backend conventions of Section~\ref{sec:webapps}. Along modality, 221 tasks rely only on text and vision (\textbf{text/vision} tasks), while 79 are \textbf{audio/video} tasks that require information embedded in temporal media on a page, such as the spoken content of a podcast or the visual content of a video.

The default action space follows the BrowserGym convention~\citep{chezellesBrowserGymEcosystemWeb2025}, with element-targeted actions (\texttt{click}, \texttt{fill}, \texttt{select\_option}, \texttt{scroll}) and navigation actions (\texttt{goto}, \texttt{go\_back}). The observation space exposes each page as both a structured accessibility tree and a screenshot, enabling evaluation of text-only and vision-language agents under the same task specifications.

To support audio/video tasks, we extend this default action space by surfacing audio and video sources as nodes in the accessibility tree and introducing four actions that operate on them (Figure~\ref{fig:overview}).
\begin{itemize}\itemsep=0pt
\item \texttt{transcribe(url, start, end)}: Returns a transcript of the audio segment from \texttt{start} to \texttt{end} (in seconds) using a local Whisper model.
\item \texttt{extract\_frames(url, n, start, end)} (vision-language agents only): Returns $n$ evenly spaced frames from the video segment between \texttt{start} and \texttt{end} (in seconds).
\item \texttt{listen(url, start, end)} (omni models only): Provides the audio segment from start to end (in seconds) as direct audio input to the model.
\item \texttt{watch(url, start, end, include\_audio)} (omni models only): Provides the video segment from start to end (in seconds) as direct video input to the model, optionally including synchronized audio.
\end{itemize}

The agent invokes these actions when it judges that the audio or video content is needed for the task. 
The resulting transcripts and extracted frames from \texttt{transcribe()} and \texttt{extract\_frames()} are returned as observations in the next step, with transcripts merged into the accessibility tree as synthetic nodes and frames returned as image content alongside the accessibility tree.
For omni models, \texttt{listen()} and \texttt{watch()} allow the videos and audio to be sent with the accessibility tree and processed by the model as a different modality.

\subsection{Evaluation}
\label{sec:evaluation}

We evaluate each trajectory with a rubric-based verifier. Every task is paired with a small set of human-written criteria, tagged as \texttt{[action]} or \texttt{[outcome]}. Action rubrics measure whether the agent performed required steps, such as searching, filtering, or submitting a form. Outcome rubrics measure whether the final state of the website or the final answer satisfies the request, and for question-answering tasks they include a reference answer. We organize the rubrics of a task into a directed acyclic graph (DAG) with one rubric node per criterion. A rubric node that depends on a failed rubric node is counted as failed. A task counts as a success only when all of its rubric nodes pass, and we report the success rate as the fraction of tasks that succeed.

Each grader decides whether a rubric is satisfied, and each rubric node uses one or more graders. There are three kinds of graders. (1) Deterministic graders query the backend state of the website, for example whether a booking exists, or look for required strings in the final answer of the agent. (2) Heuristic graders match patterns in the URLs the agent visited and the actions it took. (3) An LLM judge reads the trajectory and the rubric. When a rubric has an unambiguous expected result, such as an exact name in the answer, a deterministic grader decides whether the rubric node passes or fails on its own. Otherwise, the graders are applied in the order deterministic, heuristic, then LLM judge. If a grader finds the rubric satisfied, the rubric node passes. If not, the next grader is tried, so that a correct but unexpected solution can still be recognized. The LLM judge always comes last and makes the final decision, and it is the only grader for rubrics that cannot be verified from the backend state or the actions of the agent.

\begin{table}[t]
\centering
\small
\caption{Qualitative rubric example from the NextCloud task configuration. The judge verifies whether the agent performed actions 1 and 2, and verifies outcomes 3 and 4 at each step.}
\label{tab:nextcloud-rubric-example}
\vspace{0.5em}
\begin{tabularx}{\linewidth}{@{}c l X@{}}
\toprule
\textbf{\#} & \textbf{Type} & \textbf{Criterion} \\
\midrule
1 & Action &
The agent viewed the Project Alpha Talk history. \\
2 & Action &
The agent typed ``Client Meeting Update'' as the name for a new file. \\
3 & Outcome &
The content of the ``Client Meeting Update'' file states that the client meeting
moved to next Thursday. \\
4 & Outcome &
A file named ``Client Meeting Update'' exists in the Nextcloud file directory. \\
\bottomrule
\end{tabularx}
\end{table}

\subsection{Benchmark Construction}
\label{sec:construction}

\begin{figure*}[t]
    \centering
    \includegraphics[width=\textwidth]{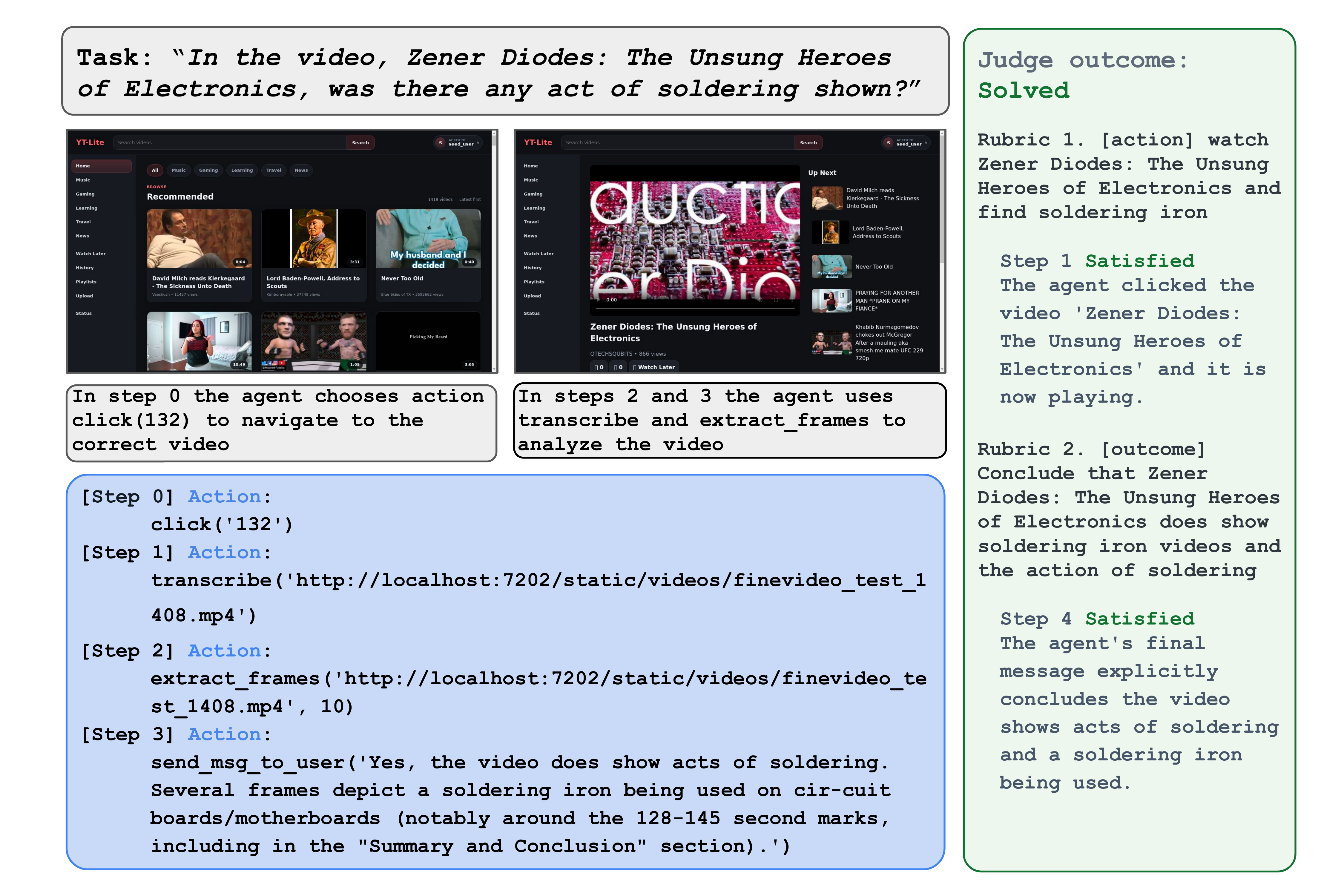}
    \caption{
        \textbf{Qualitative example of a Qwen3.6-27B agent trajectory in WebArena-Pro.} Given a single-site audio/video task on YT-Lite, namely ``\textit{In the video, Zener Diodes: The Unsung Heroes of Electronics, was there any act of soldering shown?}'', the agent first clicks the matching video thumbnail. On the video page, it invokes \texttt{transcribe()} to read the spoken content and \texttt{extract\_frames()} to sample 10 representative frames. The transcript does not mention soldering, but the extracted frames at $t=128.6$\,s and $t=144.7$\,s show a soldering iron, and the agent answers accordingly. The rubric verifier confirms a passing trajectory.
    }
    \label{fig:trajectory}
\end{figure*}

\paragraph{Mock Data and Personas.}
Each application is shipped with seeded data so that tasks operate on a realistic, non-empty state. Seed contents are application-appropriate and include documents, files, messages, calendar events, dashboards, listings, support tickets, and similar records, as well as the relations between them, such as channel memberships, file shares, and assignments. Each application is also seeded with one or more user personas, represented by credentials and pre-populated account state, so that tasks begin from an authenticated browser session. 

\paragraph{Task Creation Pipeline.}
Tasks are authored by an annotator team composed of the authors of this paper. Through a dedicated annotator interface, an annotator writes the natural-language intent, selects the relevant sites, and drafts the evaluation rubric. Each candidate then undergoes an automated triage step that flags trivial tasks by running against an LLM-based agent and discarding those solved correctly on the first attempt. Tasks that pass triage are reviewed by a second annotator on the live application, who confirms feasibility, finalizes the rubric items used for scoring, and confirms the difficulty label. Human validation is the gating step before a task enters the released benchmark, and we do not release tasks that fail any of these checks.


\section{Experiments}
\label{sec:experiments}

\subsection{Experimental Setup}

\paragraph{Environment Setup.}
We evaluate agents by running each task in a BrowserGym-compatible browser environment~\cite{chezellesBrowserGymEcosystemWeb2025}. At the beginning of an episode, the benchmark starts a fresh website container, waits for readiness, performs the required UI login, and navigates the browser to the task start URL.
The agent receives the natural-language intent, with the list of accessible websites for the task, and then acts in the browser until it calls \texttt{done()} or \texttt{send\_msg\_to\_user()}, or reaches the maximum number of steps.
Each task is limited to 50 browser actions. We report success rates using the benchmark's evaluator output.

\paragraph{Modeling.}
We evaluate representative agent architectures paired with different language, vision-language, and omni models.
Each agent has its own configuration and action space. 
For instance, GenericAgent~\cite{chezellesBrowserGymEcosystemWeb2025,drouinWorkArenaHowCapable2024} is a ReAct~\cite{yao2023react} agent that asks for chain-of-thought~\citep{wei2022chain} and an action at each step. It uses the BrowserGym~\cite{chezellesBrowserGymEcosystemWeb2025} harness to obtain observations and act on the web page.
In our setup, GenericAgent instances receive the same task intent, browser observation, and high-level action space. GenericAgent instances operate on the accessibility tree, with screenshots augmented with Set-of-Mark \cite{yang2023setofmark}. 
They also use our WebArena-Pro custom action set, which is derived from the bid action set from BrowserGym, and augmented with \verb|done()|, \verb|transcribe()|, and \verb|extract_frames()| actions, which respectively allow finishing the current task, transcribing audio segments, and extracting frames from video segments (Section~\ref{sec:tasks}). 
The action space for omni agents includes \verb|listen()| and \verb|watch()| instead, for direct modality interaction with both audio and video, respectively. 

OpenCode, Codex, and Claude Code are evaluated via the Harbor~\cite{Harbor_Framework} harness. These coding agents are provided with a Playwright MCP and an augmented media MCP to allow \verb|transcribe()| and \verb|extract_frames()| functions. Playwright MCP handles the observation extraction and action execution on the browser. Detailed setup of these agents is given in Appendix~\ref{app:coding-agents}.

We select twelve models as agent backbones: seven open-source models, including Qwen 3.5 9B, Qwen 3.6 27B, Qwen 3.8 27B, Qwen 3.8 Max Prime, Gemma4 26B A4B, Gemma4 31B~\citep{farabet_gemma_2026}, and Kimi 3.1 Pro.  And five closed-source models, including Gemini 3.1 Flash Lite, Gemini 3.1 Pro~\citep{pichai_new_2025}, Claude Opus 4.7~\citep{anthropic_introducing_nodate}, Claude Opus 5.5, and GPT 6.1 Sol. 
We evaluate Gemini3.1 Pro as both a text and an omni model. In the latter, it leverages raw audio and video files returned by the omni action space. 

For all models, we enable extended thinking---reasoning effort set to high---and set the temperature to 1.0, following the recommendations of the respective model developers. We also cap the maximum number of tokens per step at 40k~\cite{chezellesBrowserGymEcosystemWeb2025,drouinWorkArenaHowCapable2024} for GenericAgent runs.


\paragraph{Rubric-based LLM Judge.}
We use Qwen 3.6 27B as the LLM judge, the last grader of the rubric-based verifier described in Section~\ref{sec:evaluation}. We chose it over more expensive alternatives, Gemini 3.1 Pro and Kimi 2.6, because it agrees with human annotations nearly as well while being open-weight and cost-efficient (Appendix~\ref{app:judge}).

\paragraph{Metrics.}
We report task Success Rate (SR) as our primary metric, defined as the fraction of tasks for which every rubric passes under the rubric-based grader of Section~\ref{sec:evaluation}. To capture partial progress on unsuccessful tasks, we also report a partial-completion score that measures the fraction of rubric items passed per task, in Appendix~\ref{app:partial}.

\subsection{Main Results}

\begin{table*}[t]
    \centering
    \small
    \caption{Success rates (SR) of agents on WebArena-Pro at 50-step budget, broken down by domain category. Domain columns abbreviate Productivity (Prod.), Communication (Comm.), Analytics (Anlt.), Business Operations (BusOps), Design \& Media (Design), and Travel. Success requires a final reward of 1. Domain task counts are 53, 44, 16, 63, 98, and 26,
    respectively (300 overall).
    $^\dagger$Omni uses \texttt{watch()} and \texttt{listen()} functions.}
    \label{tab:results}
    \setlength{\tabcolsep}{4pt}
    \renewcommand{\arraystretch}{1.1}
    \resizebox{\linewidth}{!}{
    \begin{tabular}{llccccccc}
        \toprule
        & & \multicolumn{6}{c}{\textbf{Domains SR (\%)}} & \\
        \cmidrule(lr){3-8}
        \textbf{Agent} & \textbf{Model}
        & \textbf{Prod.} & \textbf{Comm.} & \textbf{Anlt.}
        & \textbf{BusOps} & \textbf{Design} & \textbf{Travel}
        & \textbf{Overall SR (\%)} \\
        \midrule
        \multirow{7}{*}{GenericAgent} & Qwen 3.5 9B
        & 3.8 & 9.1 & 12.5 & 3.2 & 4.1 & 3.8 & 5.0 \\
        & Qwen 3.6 27B
        & 5.7 & 22.7 & 12.5 & 7.9 & 15.3 & 15.4 & 13.0 \\
        & Qwen 3.8 27B
        & 7.5 & 15.9 & 12.5 & 14.3 & 19.4 & 23.1 & 15.7 \\
        & Gemma4 26B A4B 
        & 18.9 & 25.0 & 37.5 & 23.8 & 17.3 & 30.8 & 22.3 \\
        & Gemma4 31B
        & 18.9 & 38.6 & 37.5 & 30.2 & 30.6 & 34.6 & 30.3 \\
        & Gemini 3.1 Flash Lite
        & 15.1 & 27.3 & 25.0 & 20.6 & 25.5 & 34.6 & 23.7 \\
        & Gemini 3.1 Pro omni$^\dagger$
        & 26.4 & 27.3 & 43.8 & 20.6 & 38.8 & 26.9 & 30.3 \\
        & Claude Opus 4.7
        & 20.8 & 36.4 & 31.3 & 17.5 & 24.5 & 34.6 & 25.3 \\
        \addlinespace
        \multirow{3}{*}{OpenCode} 
        & Gemini 3.1 Pro & 41.5 & 65.9 & 50.0 & 41.3 & 50.0 & 38.5 & 48.0 \\
        & Qwen 3.8 Max Prime & 54.7 & 70.5 & 56.3 & 47.6
        & 55.1 & \textbf{53.8} & 55.7 \\
        & Kimi 3.1 Pro
        & 56.6 & 79.5 & \textbf{68.8} & 54.0 & 59.2 & 53.8 & 60.7 \\
        \addlinespace
        Codex & GPT 6.1 Sol
        & \textbf{60.4} & \textbf{81.8} & 62.5 & 49.2
        & 64.3 & \textbf{57.7} & \textbf{62.3} \\
      \addlinespace
      Claude Code & Claude Opus 5.5 & \textbf{60.4} & 72.7 & 62.5 & \textbf{57.1} & \textbf{65.3} & 50.0 & \textbf{62.3}\\
        \bottomrule
    \end{tabular}
    }
\end{table*}

Table~\ref{tab:results} reports the success rate of each agent at a 50-step budget. Both Codex with GPT 6.1 Sol and Claude Code with Claude Opus 5.5 achieves the highest overall SR at 62.3\%, followed by OpenCode with Kimi 3.1 Pro at 60.3\%, and OpenCode with Qwen 3.8 Max Prime at 55.7\%. Among GenericAgent configurations, Gemma4 31B and Gemini 3.1 Pro with omni support tie for the highest SR at 30.3\%, while the Qwen models lag well behind at  5.0\% and 15.7\%. Even the strongest configuration leaves 38\% of tasks without a recorded success. These results demonstrate remaining headroom on WebArena-Pro, while highlighting the wide variation in performance across model and agent configurations.

Performance also varies across domains, which shows the heterogeneity of the benchmark (Section~\ref{sec:webapps}). Communication yields the highest success rate for each of the five coding-agent configurations: Codex reaches 81.8\%, followed by OpenCode with Kimi at 79.5\%, Claude Code at 72.7\%, OpenCode with Qwen at 70.5\%, and OpenCode with Gemini at 65.9\%. Codex and Claude Code tie for the lead in Productivity at 60.4\%, while Claude Code leads in Design \& Media at 65.3\% and Business Operations at 57.1\%. OpenCode with Kimi leads in Analytics at 68.8\%. Travel remains challenging, with the best SR of 57.7\% achieved by Codex with GPT 6.1 Sol. No configuration leads outright in every domain, and even the strongest domain-specific results leave room for improvement.



\subsection{Site Coverage and Modality}
\label{sec:breakdowns}

In Table~\ref{tab:breakdowns_sr}, we report the success rate of each agent decomposed along two task axes, site coverage (single-site and cross-site) and modality (text/vision and audio/video). Cross-site tasks consistently yield lower success rates than single-site tasks, but the magnitude of the drop varies across agents. GenericAgent with Gemma4 31B shows the largest gap, falling from 38.2\% on single-site tasks to 13.5\% on cross-site tasks, a drop of 24.7 percentage points. Codex leads on single-site tasks at 67.2\%, but drops to 52.1\% on cross-site tasks. Claude Code achieves the highest cross-site SR at 55.2\%, compared with 65.7\% on single-site tasks, yielding the smallest gap among the five coding-agent configurations at 10.5 percentage points. These results indicate that coordinating actions across multiple applications remains a substantial challenge for current agents.

The modality breakdown reveals a more varied pattern. Within GenericAgent, audio/video tasks yield higher success rates than text/vision tasks for all models except Gemma4 26B A4B. This pattern may partly reflect the structured access provided by \texttt{transcribe()} and \texttt{extract\_frames()}, which convert audio/video content into language and image inputs that these models can process. However, this advantage does not hold consistently across coding-agent configurations. OpenCode with Qwen 3.8 Max Prime and Kimi 3.1 Pro, Codex, and Claude Code perform worse on audio/video tasks than on text/vision tasks. OpenCode with Gemini 3.1 Pro instead shows a small audio/video advantage of 1.9 percentage points. Codex and Claude Code tie for the highest SR in both modalities, achieving 64.3\% on text/vision tasks and 57.0\% on audio/video tasks. Thus, relative performance across modalities varies with the agent--model combination.



\begin{table}[t]
  \centering
  \caption{Success rate (SR) of agents on WebArena-Pro at a 50-step budget, decomposed along the two task axes of Section~\ref{sec:tasks}. The site coverage axis splits the 300 Test tasks into 204 single-site and 96 cross-site tasks,
  and the modality axis splits them into 221 text/vision and 79 audio/video tasks. Best values are bold, and second-best are underlined.}
  \vspace{0.5em}
  \label{tab:breakdowns_sr}
  \setlength{\tabcolsep}{4pt}
  \renewcommand{\arraystretch}{1.1}
  \resizebox{\linewidth}{!}{
    \small
    \begin{tabular}{llcccc}
      \toprule
      & & \multicolumn{2}{c}{\textbf{Site Coverage}}
      & \multicolumn{2}{c}{\textbf{Modality}} \\
      \cmidrule(lr){3-4}\cmidrule(lr){5-6}
      \textbf{Agent} & \textbf{Model}
      & \textbf{Single-site (\%)}
      & \textbf{Cross-site (\%)}
      & \textbf{Text/Vision (\%)}
      & \textbf{Audio/Video (\%)} \\
      \midrule
      \multirow{7}{*}{GenericAgent} & Qwen 3.5 9B
      & 5.4 & 4.2 & 5.0 & 5.1 \\
      & Qwen 3.6 27B
      & 17.2 & 4.2 & 10.9 & 19.0 \\
      & Qwen 3.8 27B
      & 20.6 & 5.2 & 12.2 & 25.3 \\
      & Gemma4 26B A4B 
      & 27.5 & 11.5 & 24.0 & 17.7 \\
      & Gemma4 31B
      & 38.2 & 13.5 & 29.9 & 31.6 \\
      & Gemini 3.1 Flash Lite
      & 29.4 & 11.5 & 22.6 & 26.6 \\
      & Gemini 3.1 Pro omni
      & 35.8 & 18.8 & 26.2 & 41.8 \\
      & Claude Opus 4.7
      & 32.8 & 9.4 & 23.1 & 31.6 \\
      \addlinespace
      \multirow{3}{*}{OpenCode} 
      & Gemini 3.1 Pro
      & 52.9 & 37.5 & 47.5 & 49.4 \\
      & Qwen 3.8 Max Prime
      & 59.8 & 46.9 & 57.5 & 50.6 \\
      & Kimi 3.1 Pro
      & \underline{66.2} & 49.0
      & \underline{63.3} & \underline{53.2} \\
      \addlinespace
      Codex & GPT 6.1 Sol
      & \textbf{67.2} & \underline{52.1}
      & \textbf{64.3} & \textbf{57.0} \\
      \addlinespace
      Claude Code & Claude Opus 5.5
      & 65.7 & \textbf{55.2}
      & \textbf{64.3} & \textbf{57.0} \\
      \bottomrule
    \end{tabular}
  }
\end{table}

\subsection{Media Tools Ablation}
\label{sec:media-ablation}

\begin{table*}[t]
\centering
\small
\setlength{\tabcolsep}{4pt}
\renewcommand{\arraystretch}{1.1}
\caption{
Media-tool ablations on the Test split with a 50-step cap.
Values report full-task SR (\%; final reward equals 1).
Results are shown for all evaluated tasks and a conservative,
post hoc media-content-dependent subset selected by auditing
task metadata, thumbnails, and grading criteria.
T/F denotes \texttt{transcribe()} and \texttt{extract\_frames()};
W/L denotes \texttt{watch()} and \texttt{listen()}.
}
\label{tab:media-tool-ablations-all}
\resizebox{\linewidth}{!}{
\begin{tabular}{llrrrrrr}
\toprule
& & \multicolumn{3}{c}{\textbf{All Tasks}}
  & \multicolumn{3}{c}{\textbf{Media-Content-Dependent Subset}} \\
\cmidrule(lr){3-5}
\cmidrule(lr){6-8}
\textbf{Model} & \textbf{Configuration}
& \textbf{AcidWave} & \textbf{YT-Lite} & \textbf{Overall}
& \textbf{Audio} & \textbf{Video} & \textbf{Overall} \\
& &
($n=35$) & ($n=44$) & ($n=79$)
& ($n=20$) & ($n=23$) & ($n=43$) \\
\midrule
Gemini3.1 Pro
 & No media tools & 11.4 & 20.5 & 16.5 & 0.0  & 4.3  & 2.3 \\
 & T/F only       & 22.9 & 50.0 & 38.0 & 5.0  & 52.2 & 30.2 \\
 & W/L only       & 25.7 & 54.5 & 41.8 & 10.0 & 56.5 & 34.9 \\
\midrule
Qwen 3.6 27B
 & No media tools & 5.7 & 2.3  & 3.8  & 0.0 & 0.0  & 0.0 \\
 & T/F only       & 8.6 & 27.3 & 19.0 & 0.0 & 30.4 & 16.3 \\
 \midrule
 Opus 5.5 & No media tools & 17.1 & 34.1 & 26.6 & 5.0 & 21.7 & 14.0 \\
 & T/F only & 31.4 & 77.3 & 57.0 & 15.0 & 78.3 & 48.8 \\
\bottomrule
\end{tabular}
}
\end{table*}

In this section, we report an ablation study on how much media access contributes to agent performance. We compare three configurations. (1) No media tools, where the agent cannot access audio or video content. (2) Tool-based media access (T/F only), where the agent calls \texttt{transcribe()} to obtain transcripts of audio and \texttt{extract\_frames()} to obtain frames of videos. (3) Native omni access (W/L only), where the omni model Gemini 3.1 Pro receives the audio and video content directly as input. We evaluate all 79 tasks on AcidWave and YT-Lite, the two websites centered on audio and video. We also report results on a subset of 43 tasks that cannot be solved from titles, metadata, or thumbnails alone. 
Table~\ref{tab:media-tool-ablations-all} shows the results. 
On this subset, removing the media tools drops Gemini 3.1 Pro and Qwen 3.6 27B to 0\% on audio tasks; Gemini 3.1 Pro solves a single video task, which we attribute to a correct guess. Claude Code with Opus 5.5 retains 14.0\% without media tools but rises to 48.8\% with them.

\section{Discussion}
\label{sec:discussion}

\paragraph{Costs.}
The cost of each agent run depends on its architecture, harness, and underlying model. We report recorded API costs for a 300-task evaluation, including retries and recovery runs but excluding hosting infrastructure. GenericAgent costs approximately USD 800 with Claude Opus 4.7. For the coding agents, recorded costs including agent inference are USD 386 for Claude Code with Opus 5.5, USD 166 for Codex with GPT 6.1 Sol, USD 689 for OpenCode with Qwen 3.8 Max Prime, and USD 340 for OpenCode with Kimi 3.1 Pro. Judging with Qwen 3.6 27B through OpenRouter accounts for less than USD 4 per 300-task evaluation in these coding-agent runs.


\paragraph{Future work.}
The same self-hosted infrastructure that supports parallel evaluation can also serve as a training environment, where the rubric-based verifier provides a per-task reward signal for reinforcement learning or rejection sampling. The controlled environment also supports robustness studies that perturb the underlying web pages to test resilience to UI drift and prompt injection. Finally, expanding the action space to richer temporal modalities such as live voice and streaming video would let the benchmark track agents that move beyond static page interaction.
\section{Conclusion}

We presented WebArena-Pro, a reproducible benchmark of 300 human-validated tasks across 20 self-hosted web applications spanning six domain categories. It treats audio and video as core observations alongside text and vision, and runs each task in container isolation for reproducible parallel evaluation. 
Across twelve model backbones and four agent harnesses, the strongest agent, Claude Code with Opus 5.5 and Codex with GPT 6.1 Sol, reach only 62.3\% success at a 50-step budget, with cross-site tasks remaining harder than single-site ones, and media-access ablations demonstrate the importance of accessing audio and video content. These results indicate considerable headroom for progress on heterogeneous, multimodal web agents.

\section*{Limitations}
We note the following limitations to our work.

\paragraph{Limited Temporal Interaction.}
Our action-space extension covers only prerecorded audio segments and discrete video frames, leaving richer temporal interaction modalities such as voice dialogue and live video streams to future work.

\paragraph{Evaluation Comparability and Reliability.}
Our comparisons evaluate particular combinations of models, harnesses, and tool configurations. Performance differences cannot generally be attributed to model capability or harness design alone. A shared 50-step budget also does not equalize computational resources: agents differ in token consumption, native tool use, observation frequency, and latency. Reported costs depend on provider pricing and cache utilization. Controlled comparisons across harnesses and repeated runs would strengthen conclusions about efficiency and reliability.

\section*{Acknowledgements}
Xing Han Lù acknowledges the support of the Natural Sciences and Engineering Research Council of Canada (NSERC) [funding reference no. 579403].
Siva Reddy is supported by the Canada CIFAR AI Chair program, an NSERC Discovery Grant, and a Sloan Fellowship. This project is also supported by the ServiceNow–Mitacs Accelerate program. Gunhee Kim, Fatemeh Pesaran Zadeh, Dongjin Kang, and Seyeon Choi are supported by the IITP(Institute of Information \& Communications Technology Planning \& Evaluation)-ITRC(Information Technology Research Center) grant funded by the Korea government(Ministry of Science and ICT)(IITP-2025-RS-2024-00437633), Institute of Information \& Communications Technology Planning \& Evaluation(IITP) grant funded by the Korea government(MSIT) [No. RS-2025-25442338, AI star Fellowship Support Program(Seoul National Univ.); No. RS-2021-II211343, Artificial Intelligence Graduate School Program (Seoul National University)], and Basic Science Research Program through the National Research Foundation of Korea(NRF) funded by the Ministry of Education(RS-2023-00274280).



\bibliographystyle{abbrv}
\bibliography{bibliography}

@inproceedings{
hattami2025webarena,
title={WebArena Verified: Reliable Evaluation for Web Agents},
author={Amine El hattami and Megh Thakkar and Nicolas Chapados and Christopher Pal},
booktitle={Workshop on Scaling Environments for Agents},
year={2025},
url={https://openreview.net/forum?id=94tlGxmqkN}
}

@inproceedings{yao2023react,
  title     = {{ReAct}: Synergizing Reasoning and Acting in Language Models},
  author    = {Yao, Shunyu and Zhao, Jeffrey and Yu, Dian and Du, Nan and Shafran, Izhak and Narasimhan, Karthik and Cao, Yuan},
  booktitle = {International Conference on Learning Representations (ICLR)},
  year      = {2023},
  url       = {https://arxiv.org/abs/2210.03629}
}

@software{Harbor_Framework,
  author = {{Harbor Framework Team}},
  title  = {{Harbor: A framework for evaluating and optimizing agents and models in container environments}},
  year   = {2026},
  doi    = {10.5281/zenodo.20953922},
  url    = {https://doi.org/10.5281/zenodo.20953922}
}

@misc{pichai_new_2025,
    title    = {A new era of intelligence with {Gemini} 3},
    author   = {Pichai, Sundar and Hassabis, Demis and Kavukcuoglu, Koray},
    journal  = {Google},
    month    = {November},
    year     = {2025},
    url      = {https://blog.google/products-and-platforms/products/gemini/gemini-3/},
    urldate  = {2026-05-06},
    language = {en-us}
}

@misc{farabet_gemma_2026,
    title      = {{Gemma 4}: {Byte} for byte, the most capable open models},
    shorttitle = {{Gemma 4}},
    author     = {Farabet, Clement and Lacombe, Olivier},
    journal    = {Google},
    month      = {April},
    year       = {2026},
    url        = {https://blog.google/innovation-and-ai/technology/developers-tools/gemma-4/},
    urldate    = {2026-05-06},
    language   = {en-us}
}

@misc{anthropic_introducing_nodate,
    title    = {Introducing {Claude} {Opus} 4.7},
    author   = {{Anthropic}},
    url      = {https://www.anthropic.com/news/claude-opus-4-7},
    urldate  = {2026-05-06},
    language = {en}
}

@article{xue2025illusion,
  title={An illusion of progress? assessing the current state of web agents},
  author={Xue, Tianci and Qi, Weijian and Shi, Tianneng and Song, Chan Hee and Gou, Boyu and Song, Dawn and Sun, Huan and Su, Yu},
  journal={arXiv preprint arXiv:2504.01382},
  year={2025}
}

@inproceedings{osworld2024neurips,
 author = {Xie, Tianbao and Zhang, Danyang and Chen, Jixuan and Li, Xiaochuan and Zhao, Siheng and Cao, Ruisheng and Hua, Toh Jing and Cheng, Zhoujun and Shin, Dongchan and Lei, Fangyu and Liu, Yitao and Xu, Yiheng and Zhou, Shuyan and Savarese, Silvio and Xiong, Caiming and Zhong, Victor and Yu, Tao},
 booktitle = {Advances in Neural Information Processing Systems},
 doi = {10.52202/079017-1650},
 editor = {A. Globerson and L. Mackey and D. Belgrave and A. Fan and U. Paquet and J. Tomczak and C. Zhang},
 pages = {52040--52094},
 publisher = {Curran Associates, Inc.},
 title = {OSWorld: Benchmarking Multimodal Agents for Open-Ended Tasks in Real Computer Environments},
 url = {https://proceedings.neurips.cc/paper_files/paper/2024/file/5d413e48f84dc61244b6be550f1cd8f5-Paper-Datasets_and_Benchmarks_Track.pdf},
 volume = {37},
 year = {2024}
}

@misc{miyai2025webchorearenaevaluatingwebbrowsing,
      title={WebChoreArena: Evaluating Web Browsing Agents on Realistic Tedious Web Tasks}, 
      author={Atsuyuki Miyai and Zaiying Zhao and Kazuki Egashira and Atsuki Sato and Tatsumi Sunada and Shota Onohara and Hiromasa Yamanishi and Mashiro Toyooka and Kunato Nishina and Ryoma Maeda and Kiyoharu Aizawa and Toshihiko Yamasaki},
      year={2025},
      eprint={2506.01952},
      archivePrefix={arXiv},
      primaryClass={cs.CL},
      url={https://arxiv.org/abs/2506.01952}, 
}

@inproceedings{
wei2022chain,
title={Chain of Thought Prompting Elicits Reasoning in Large Language Models},
author={Jason Wei and Xuezhi Wang and Dale Schuurmans and Maarten Bosma and brian ichter and Fei Xia and Ed H. Chi and Quoc V Le and Denny Zhou},
booktitle={Advances in Neural Information Processing Systems},
editor={Alice H. Oh and Alekh Agarwal and Danielle Belgrave and Kyunghyun Cho},
year={2022},
url={https://openreview.net/forum?id=_VjQlMeSB_J}
}

@misc{xu2025theagentcompanybenchmarkingllmagents,
      title={TheAgentCompany: Benchmarking LLM Agents on Consequential Real World Tasks}, 
      author={Frank F. Xu and Yufan Song and Boxuan Li and Yuxuan Tang and Kritanjali Jain and Mengxue Bao and Zora Z. Wang and Xuhui Zhou and Zhitong Guo and Murong Cao and Mingyang Yang and Hao Yang Lu and Amaad Martin and Zhe Su and Leander Maben and Raj Mehta and Wayne Chi and Lawrence Jang and Yiqing Xie and Shuyan Zhou and Graham Neubig},
      year={2025},
      eprint={2412.14161},
      archivePrefix={arXiv},
      primaryClass={cs.CL},
      url={https://arxiv.org/abs/2412.14161}, 
}

@misc{abuelsaadAgentEAutonomousWeb2024,
  title = {Agent-{{E}}: {{From Autonomous Web Navigation}} to {{Foundational Design Principles}} in {{Agentic Systems}}},
  shorttitle = {Agent-{{E}}},
  author = {Abuelsaad, Tamer and Akkil, Deepak and Dey, Prasenjit and Jagmohan, Ashish and Vempaty, Aditya and Kokku, Ravi},
  year = {2024},
  month = jul,
  number = {arXiv:2407.13032},
  eprint = {2407.13032},
  primaryclass = {cs},
  publisher = {arXiv},
  doi = {10.48550/arXiv.2407.13032},
  urldate = {2025-01-24},
  archiveprefix = {arXiv}
}

@misc{boisvertWorkArenaCompositionalPlanning2024,
  title = {{{WorkArena}}++: {{Towards Compositional Planning}} and {{Reasoning-based Common Knowledge Work Tasks}}},
  shorttitle = {{{WorkArena}}++},
  author = {Boisvert, L{\'e}o and Thakkar, Megh and Gasse, Maxime and Caccia, Massimo and Chezelles, Thibault Le Sellier De and Cappart, Quentin and Chapados, Nicolas and Lacoste, Alexandre and Drouin, Alexandre},
  year = {2024},
  month = jul,
  number = {arXiv:2407.05291},
  eprint = {2407.05291},
  publisher = {arXiv},
  doi = {10.48550/arXiv.2407.05291},
  urldate = {2024-11-26},
  archiveprefix = {arXiv}
}

@misc{chezellesBrowserGymEcosystemWeb2025,
  title = {The {{BrowserGym Ecosystem}} for {{Web Agent Research}}},
  author = {Chezelles, Thibault Le Sellier De and Gasse, Maxime and Drouin, Alexandre and Caccia, Massimo and Boisvert, L{\'e}o and Thakkar, Megh and Marty, Tom and Assouel, Rim and Shayegan, Sahar Omidi and Jang, Lawrence Keunho and L{\`u}, Xing Han and Yoran, Ori and Kong, Dehan and Xu, Frank F. and Reddy, Siva and Cappart, Quentin and Neubig, Graham and Salakhutdinov, Ruslan and Chapados, Nicolas and Lacoste, Alexandre},
  year = {2025},
  month = feb,
  number = {arXiv:2412.05467},
  eprint = {2412.05467},
  primaryclass = {cs},
  publisher = {arXiv},
  doi = {10.48550/arXiv.2412.05467},
  urldate = {2025-04-25},
  archiveprefix = {arXiv}
}

@misc{dengMind2WebGeneralistAgent2023,
  title = {{{Mind2Web}}: {{Towards}} a {{Generalist Agent}} for the {{Web}}},
  shorttitle = {{{Mind2Web}}},
  author = {Deng, Xiang and Gu, Yu and Zheng, Boyuan and Chen, Shijie and Stevens, Samuel and Wang, Boshi and Sun, Huan and Su, Yu},
  year = {2023},
  month = jun,
  number = {arXiv:2306.06070},
  eprint = {2306.06070},
  primaryclass = {cs},
  publisher = {arXiv},
  urldate = {2023-10-31},
  archiveprefix = {arXiv}
}

@misc{drouinWorkArenaHowCapable2024,
  title = {{{WorkArena}}: {{How Capable Are Web Agents}} at {{Solving Common Knowledge Work Tasks}}?},
  shorttitle = {{{WorkArena}}},
  author = {Drouin, Alexandre and Gasse, Maxime and Caccia, Massimo and Laradji, Issam H. and Del Verme, Manuel and Marty, Tom and Boisvert, L{\'e}o and Thakkar, Megh and Cappart, Quentin and Vazquez, David and Chapados, Nicolas and Lacoste, Alexandre},
  year = {2024},
  month = mar,
  number = {arXiv:2403.07718},
  eprint = {2403.07718},
  primaryclass = {cs},
  publisher = {arXiv},
  urldate = {2024-04-20},
  archiveprefix = {arXiv}
}

@misc{gurRealWorldWebAgentPlanning2023,
  title = {A {{Real-World WebAgent}} with {{Planning}}, {{Long Context Understanding}}, and {{Program Synthesis}}},
  author = {Gur, Izzeddin and Furuta, Hiroki and Huang, Austin and Safdari, Mustafa and Matsuo, Yutaka and Eck, Douglas and Faust, Aleksandra},
  year = {2023},
  month = oct,
  number = {arXiv:2307.12856},
  eprint = {2307.12856},
  primaryclass = {cs},
  publisher = {arXiv},
  urldate = {2023-10-31},
  archiveprefix = {arXiv}
}

@misc{heWebVoyagerBuildingEndtoEnd2024,
  title = {{{WebVoyager}}: {{Building}} an {{End-to-End Web Agent}} with {{Large Multimodal Models}}},
  shorttitle = {{{WebVoyager}}},
  author = {He, Hongliang and Yao, Wenlin and Ma, Kaixin and Yu, Wenhao and Dai, Yong and Zhang, Hongming and Lan, Zhenzhong and Yu, Dong},
  year = {2024},
  month = feb,
  number = {arXiv:2401.13919},
  eprint = {2401.13919},
  primaryclass = {cs},
  publisher = {arXiv},
  urldate = {2024-03-12},
  archiveprefix = {arXiv}
}

@misc{kohVisualWebArenaEvaluatingMultimodal2024,
  title = {{{VisualWebArena}}: {{Evaluating Multimodal Agents}} on {{Realistic Visual Web Tasks}}},
  shorttitle = {{{VisualWebArena}}},
  author = {Koh, Jing Yu and Lo, Robert and Jang, Lawrence and Duvvur, Vikram and Lim, Ming Chong and Huang, Po-Yu and Neubig, Graham and Zhou, Shuyan and Salakhutdinov, Ruslan and Fried, Daniel},
  year = {2024},
  month = jan,
  journal = {arXiv.org},
  urldate = {2025-04-23},
  howpublished = {https://arxiv.org/abs/2401.13649v2},
  langid = {english}
}

@inproceedings{jang2025videowebarena,
  title     = {VideoWebArena: Evaluating Long Context Multimodal Agents with Video Understanding Web Tasks},
  author    = {Jang, Lawrence and Li, Yinheng and Zhao, Dan and Ding, Charles and Lin, Justin and Liang, Paul Pu and Bonatti, Rogerio and Koishida, Kazuhito},
  booktitle = {The Thirteenth International Conference on Learning Representations (ICLR)},
  year      = {2025}
}

@article{yang2023setofmark,
      title={Set-of-Mark Prompting Unleashes Extraordinary Visual Grounding in GPT-4V}, 
      author={Jianwei Yang and Hao Zhang and Feng Li and Xueyan Zou and Chunyuan Li and Jianfeng Gao},
      journal={arXiv preprint arXiv:2310.11441},
      year={2023},
}

@misc{zhouWebArenaRealisticWeb2023,
  title = {{{WebArena}}: {{A Realistic Web Environment}} for {{Building Autonomous Agents}}},
  shorttitle = {{{WebArena}}},
  author = {Zhou, Shuyan and Xu, Frank F. and Zhu, Hao and Zhou, Xuhui and Lo, Robert and Sridhar, Abishek and Cheng, Xianyi and Ou, Tianyue and Bisk, Yonatan and Fried, Daniel and Alon, Uri and Neubig, Graham},
  year = {2023},
  month = oct,
  number = {arXiv:2307.13854},
  eprint = {2307.13854},
  primaryclass = {cs},
  publisher = {arXiv},
  doi = {10.48550/arXiv.2307.13854},
  urldate = {2023-11-28},
  archiveprefix = {arXiv}
}

@misc{aggarwal2026gymanythingturnsoftwareagent,
      title={Gym-Anything: Turn any Software into an Agent Environment}, 
      author={Pranjal Aggarwal and Graham Neubig and Sean Welleck},
      year={2026},
      eprint={2604.06126},
      archivePrefix={arXiv},
      primaryClass={cs.LG},
      url={https://arxiv.org/abs/2604.06126}, 
}


\appendix
\clearpage
\section{Coding Agents}
\label{app:coding-agents}

In this section, we detail the setup of coding agents, especially describing their action space. 
Table~\ref{tab:coding-agent-action-spaces} shows the tools available for the agents at runtime. These actions are available through Playwright MCP for basic navigation functions and WebArena-Pro MCP, which adds media functions and \verb|webarena_submit()|. Browser navigation is restricted to the selected task websites.

\begin{figure}[p]
    \centering
    \includegraphics[width=\linewidth]{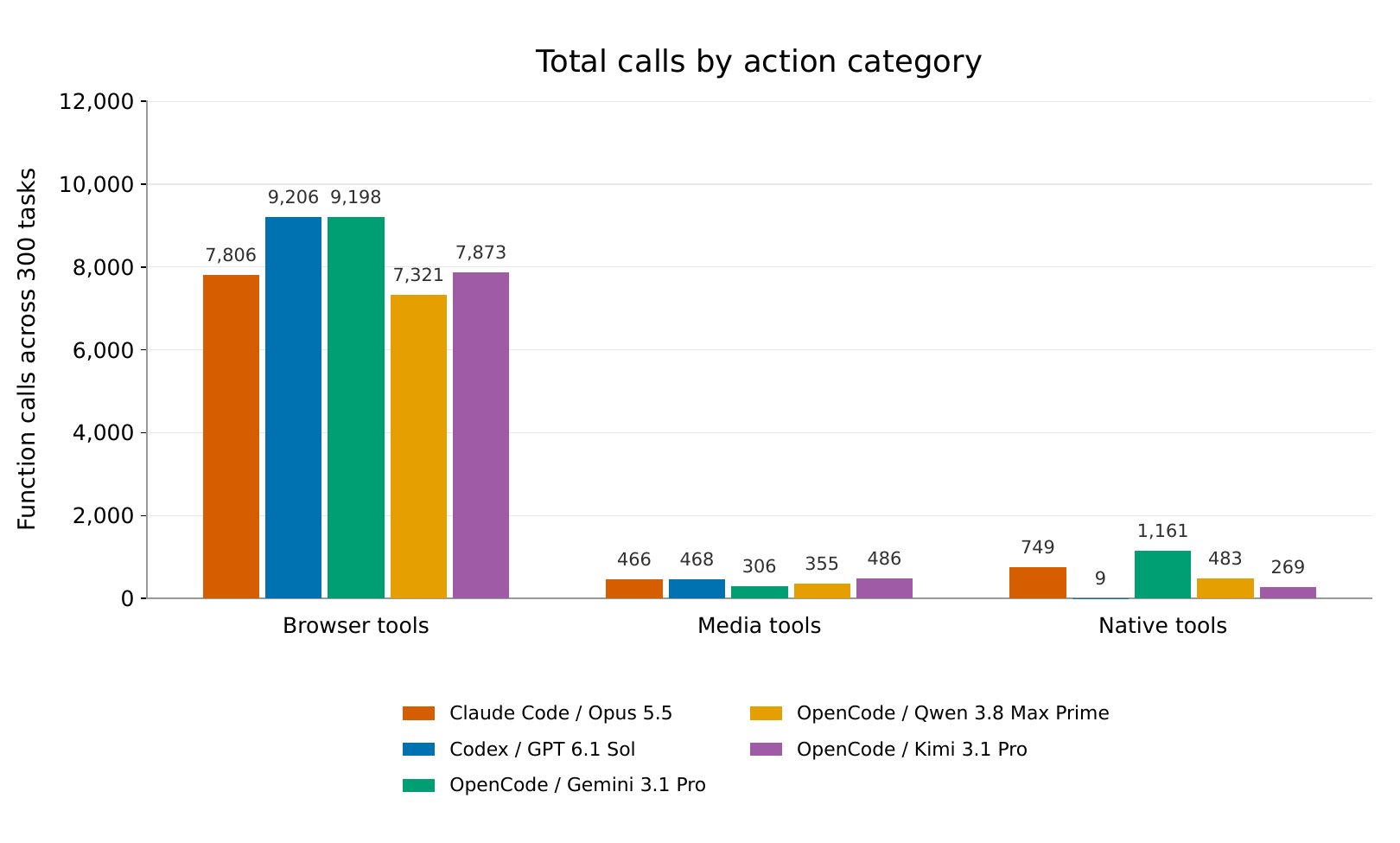}
    \caption{Tool-call frequencies by action category on the WebArena-Pro Test split (300 tasks per coding agent). Bars show total recorded browser, media, and native tool invocations, including observation calls. Submission calls and Codex's \texttt{exec}/\texttt{wait} orchestration wrappers are excluded.}
    \label{fig:coding-agents-call-freq}
\end{figure}

\begin{figure}[p]
    \centering
    \includegraphics[width=\linewidth]{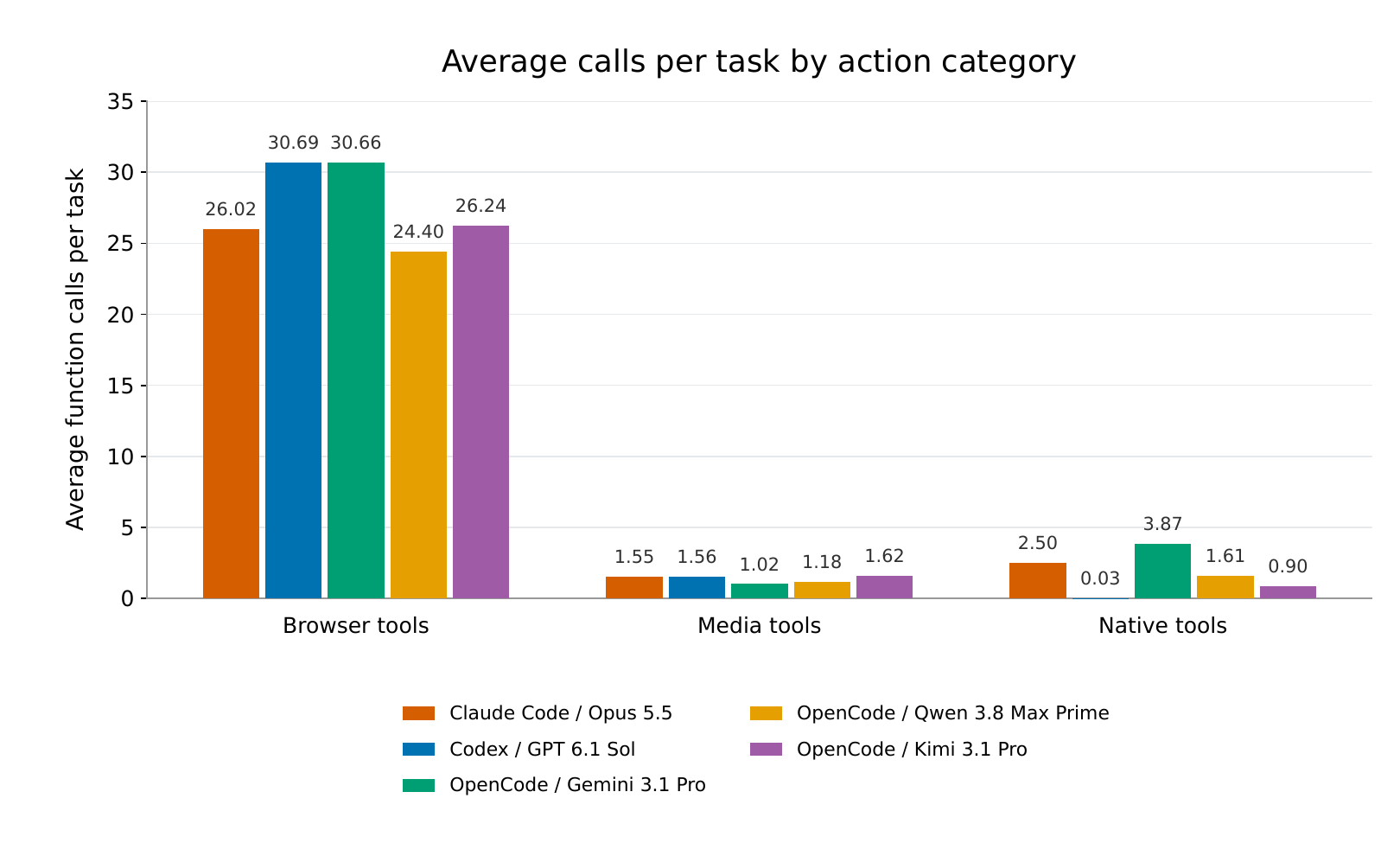}
    \caption{Average tool-call frequencies by action category per task. Bars show the average recorded browser, media, and native tool invocations, including observation calls. Submission calls and Codex's \texttt{exec}/\texttt{wait} orchestration wrappers are excluded.}
    \label{fig:coding-agents-call-freq-avg}
\end{figure}

Figure~\ref{fig:coding-agents-call-freq} shows the total tool-call frequencies over the full benchmark by tool category among: browser tools, media tools, and native tools.
Detailed tool-call frequencies are given in Table~\ref{tab:coding-agent-call-frequencies}.
The figure shows that all coding agents heavily relied on browser tools to solve WebArena-Pro tasks. Media tools were also important for audio and video interaction, as shown in Section~\ref{sec:media-ablation}. Finally, we see that Claude Code powered by Opus 5.5 and OpenCode powered by Gemini 3.1 Pro have both used their native tools more frequently than Kimi 3.1 Pro, Qwen 3.8 Max Prime and GPT 6.1 Sol agents. However, Codex powered by GPT 6.1 Sol had high usage of its own orchestration tools (\verb|exec| and \verb|wait|).
The average number of tool calls per task for each agent is: 31.1 for Claude Code, 33.3 for Codex, 36.5 for OpenCode with Gemini, 28.1 for OpenCode with Qwen 3.8 Max Prime, and 29.8 for OpenCode with Kimi (see Table~\ref{tab:coding-agent-call-frequencies}). Detailed average per tool category is given in Figure~\ref{fig:coding-agents-call-freq-avg}.

\begin{table*}[h]
    \centering
    \small
    \caption{Configured action spaces of the three coding agents.
    Availability refers to the evaluated runs.}
    \label{tab:coding-agent-action-spaces}
    \begin{tabular}{lccc}
        \toprule
        \textbf{Tool / capability}
        & \textbf{Claude Code}
        & \textbf{Codex}
        & \textbf{OpenCode} \\
        \midrule
        \multicolumn{4}{@{}l}{\textit{Browser tools}} \\
        Snapshot, screenshot, and page search
        & Yes & Yes & Yes \\
        Navigate, go back, and manage tabs
        & Yes & Yes & Yes \\
        Element-based click, hover, and drag
        & Yes & Yes & Yes \\
        Type, fill forms, select options, and press keys
        & Yes & Yes & Yes \\
        Handle dialogs, wait, and upload staged files
        & Yes & Yes & Yes \\
        Coordinate-based mouse actions
        & Yes & Yes & Yes \\
        Arbitrary browser JavaScript execution
        & No & No & No \\
        \addlinespace

        \multicolumn{4}{@{}l}{\textit{Media and submission tools}} \\
        \texttt{media\_transcribe}
        & Yes & Yes & Yes \\
        \texttt{media\_extract\_frames}
        & Yes & Yes & Yes \\
        Native audio/video input
        & No & No & No \\
        \texttt{webarena\_submit}
        & Yes & Yes & Yes \\
        \addlinespace

        \multicolumn{4}{@{}l}{\textit{Native coding-agent tools}} \\
        Terminal / shell execution
        & Yes & Yes & Yes \\
        Local file reading, writing, and editing
        & Yes & Yes & Yes \\
        Subagent tools
        & Yes & Yes & Yes \\
        Built-in web search
        & Disabled & Disabled & Disabled \\
        \bottomrule
    \end{tabular}
\end{table*}

\section{Additional Results}

\subsection{Additional Models}

Table~\ref{tab:results-additional} shows the SR of GenericAgent powered by Qwen 3.6 27B and Gemini 3.1 Pro on the Test set (300 tasks).

\begin{table*}[t]
    \centering
    \small
    \caption{Success rates (SR) of agents on WebArena-Pro at 50-step budget, broken down by domain category. Domain columns abbreviate Productivity (Prod.), Communication (Comm.), Analytics (Anlt.), Business Operations (BusOps), Design \& Media (Design), and Travel. Success requires a final reward of 1. Domain task counts are 53, 44, 16, 63, 98, and 26,
    respectively (300 overall).}
    \label{tab:results-additional}
    \setlength{\tabcolsep}{4pt}
    \renewcommand{\arraystretch}{1.1}
    \resizebox{\linewidth}{!}{
    \begin{tabular}{llccccccc}
        \toprule
        & & \multicolumn{6}{c}{\textbf{Domains SR (\%)}} & \\
        \cmidrule(lr){3-8}
        \textbf{Agent} & \textbf{Model}
        & \textbf{Prod.} & \textbf{Comm.} & \textbf{Anlt.}
        & \textbf{BusOps} & \textbf{Design} & \textbf{Travel}
        & \textbf{Overall SR (\%)} \\
        \midrule
        \multirow{2}{*}{GenericAgent} 
        & Qwen 3.6 27B
        & 5.7 & 22.7 & 12.5 & 7.9 & 15.3 & 15.4 & 13.0 \\
        & Gemini 3.1 Pro
        & 28.3 & 27.3 & 43.8 & 20.6 & 33.7 & 30.8 & \textbf{29.3} \\
        \bottomrule
    \end{tabular}
    }
\end{table*}

\subsection{Partial Completion Score}
\label{app:partial}

Success rate gives credit only when every rubric of a task is satisfied. To also measure partial progress, we compute a partial completion score. The rubrics of each task form a directed acyclic graph (DAG) with one node per rubric and optional prerequisite dependencies. If a prerequisite fails, the dependent node is counted as failed. For a task with $n$ rubric nodes, the partial completion score is $S = \frac{1}{n}\sum_{i=1}^{n} b_i$, where $b_i = 1$ if rubric node $i$ passes and $b_i = 0$ otherwise. A task counts as a success only when $S = 1$, so the partial completion score never changes the success rate. Table~\ref{tab:partial} reports the average partial completion score over the benchmark.
\begin{table*}[p]
    \centering
    \small
    \caption{Partial completion scores of agents on the WebArena-Pro test set (300 tasks) with a 50-step budget, broken down by domain. Scores are 100 times the mean final task reward. Domain columns abbreviate Productivity (Prod.), Communication (Comm.), Analytics (Anlt.), Business Operations (BusOps), Design \& Media (Design), and Travel. Best and second-best overall values are bold and underlined, respectively.}
    \label{tab:partial}
    \setlength{\tabcolsep}{4pt}
    \renewcommand{\arraystretch}{1.1}
    \begin{tabular}{llccccccc}
        \toprule
        & & \multicolumn{6}{c}{\textbf{Domains (\%)}} & \\
        \cmidrule(lr){3-8}
        \textbf{Agent} & \textbf{Model}
        & \textbf{Prod.} & \textbf{Comm.} & \textbf{Anlt.}
        & \textbf{BusOps} & \textbf{Design} & \textbf{Travel}
        & \textbf{Overall (\%)} \\
        \midrule
        \multirow{8}{*}{GenericAgent} & Qwen 3.5 9B
        & 17.4 & 25.7 & 21.0 & 9.1 & 18.9 & 14.6 & 17.3 \\
        & Qwen 3.6 27B
        & 17.6 & 33.6 & 31.9 & 17.3 & 24.7 & 38.5 & 24.8 \\
        & Qwen 3.8 27B
        & 20.6 & 31.2 & 30.1 & 23.7 & 33.2 & 44.2 & 29.5 \\
        & Gemma4 26B A4B
        & 38.2 & 39.9 & 48.8 & 33.8 & 29.9 & 52.5 & 36.6 \\
        & Gemma4 31B
        & 42.3 & 48.5 & 61.3 & 43.5 & 45.1 & 49.2 & 46.0 \\
        & Gemini 3.1 Flash Lite
        & 30.7 & 38.5 & 44.4 & 29.5 & 35.2 & 50.8 & 35.5 \\
        & Gemini 3.1 Pro omni
        & 50.9 & 44.5 & 62.8 & 37.4 & 51.9 & 55.0 & 48.5 \\
        & Claude Opus 4.7
        & 29.7 & 46.6 & 46.0 & 24.3 & 34.1 & 53.7 & 35.4 \\
        \addlinespace
        \multirow{3}{*}{OpenCode}
        & Gemini 3.1 Pro
        & 62.6 & 76.9 & 71.1 & 57.4 & 60.7 & 67.1 & 63.8 \\
        & Qwen 3.8 Max Prime
        & 72.3 & 78.8 & 73.2 & 64.1 & 64.3 & 71.8 & 68.9 \\
        & Kimi 3.1 Pro
        & 73.2 & 89.0 & 80.2 & 72.2 & 68.2 & 70.0 & \underline{73.8} \\
        \addlinespace
        Codex & GPT 6.1 Sol
        & 77.7 & 89.0 & 88.3 & 66.6 & 71.2 & 72.4 & \textbf{75.0} \\
        \addlinespace
        Claude Code & Claude Opus 5.5
        & 74.8 & 84.3 & 79.5 & 69.7 & 70.5 & 69.2 & 73.5 \\
        \bottomrule
    \end{tabular}
\end{table*}

\subsection{Results on the Dev Set}
\label{app:dev}

The development (dev) set is a separate set of 100 tasks, kept apart from the 300 test tasks. Its tasks are easier to solve than the test tasks, so agents reach much higher scores. Table~\ref{tab:results_dev} reports both the success rate and the partial completion score on the dev set.

\begin{table*}[p]
    \centering
    \small
    \caption{Success rates and partial completion scores of agents (variants of GenericAgent) on the WebArena-Pro dev set (100 tasks) with a 50-step budget. Domain columns abbreviate Productivity (Prod.), Communication (Comm.), Analytics (Anlt.), Business Operations (BusOps), Design \& Media (Design), and Travel.}
    \label{tab:results_dev}
    \setlength{\tabcolsep}{4pt}
    \renewcommand{\arraystretch}{1.1}
    \begin{tabular}{lccccccc}
        \toprule
        & \multicolumn{6}{c}{\textbf{Domains (\%)}} & \\
        \cmidrule(lr){2-7}
        \textbf{Model}
        & \textbf{Prod.} & \textbf{Comm.} & \textbf{Anlt.}
        & \textbf{BusOps} & \textbf{Design} & \textbf{Travel}
        & \textbf{Overall (\%)} \\
        \midrule
        \multicolumn{8}{l}{\textit{Success rate}} \\
        \midrule
        Qwen 3.5 9B
        & 10.0 & 26.7 & 57.1 & 13.3 & 37.5 & 7.1 & 28.0 \\
        Qwen 3.6 27B
        & 40.0 & 53.3 & 71.4 & 40.0 & 53.1 & 64.3 & 54.0 \\
        Gemma4 26B A4B
        & 20.0 & 73.3 & 100.0 & 66.7 & 46.9 & 57.1 & 60.0 \\
        Gemma4 31B
        & 30.0 & 80.0 & 100.0 & 66.7 & 71.9 & 85.7 & 74.0 \\
        Gemini 3.1 Flash Lite
        & 70.0 & 66.7 & 92.9 & 80.0 & 71.9 & 71.4 & \underline{75.0} \\
        Gemini 3.1 Pro
        & 70.0 & 73.3 & 100.0 & 66.7 & 75.0 & 78.6 & \textbf{77.0} \\
        Claude Opus 4.7
        & 60.0 & 53.3 & 100.0 & 53.3 & 78.1 & 71.4 & 71.0 \\
        \midrule
        \multicolumn{8}{l}{\textit{Partial completion score}} \\
        \midrule
        Qwen 3.5 9B
        & 21.7 & 42.0 & 57.1 & 13.3 & 47.7 & 19.9 & 36.5 \\
        Qwen 3.6 27B
        & 45.0 & 58.9 & 71.4 & 45.6 & 59.4 & 72.5 & 59.3 \\
        Gemma4 26B A4B
        & 28.3 & 73.3 & 100.0 & 70.5 & 59.4 & 67.1 & 66.8 \\
        Gemma4 31B
        & 35.0 & 80.0 & 100.0 & 72.7 & 76.6 & 88.6 & 77.3 \\
        Gemini 3.1 Flash Lite
        & 70.0 & 75.3 & 92.9 & 80.0 & 79.7 & 77.9 & \underline{79.7} \\
        Gemini 3.1 Pro
        & 70.0 & 73.3 & 100.0 & 72.2 & 82.8 & 83.2 & \textbf{81.0} \\
        Claude Opus 4.7
        & 63.3 & 56.7 & 100.0 & 53.3 & 82.8 & 82.0 & 74.8 \\
        \bottomrule
    \end{tabular}
\end{table*}




\section{Failure Modes Analysis}

\subsection{GenericAgent Failure Modes}

Across the eight models, failures involve incomplete workflows, unsuccessful navigation, incorrect selections, and premature termination. These patterns can overlap: an agent may repeatedly search for the wrong entity, exhaust its action budget, and leave a required action unfinished. 

\begin{figure}[t]
    \centering
    \includegraphics[width=\linewidth]{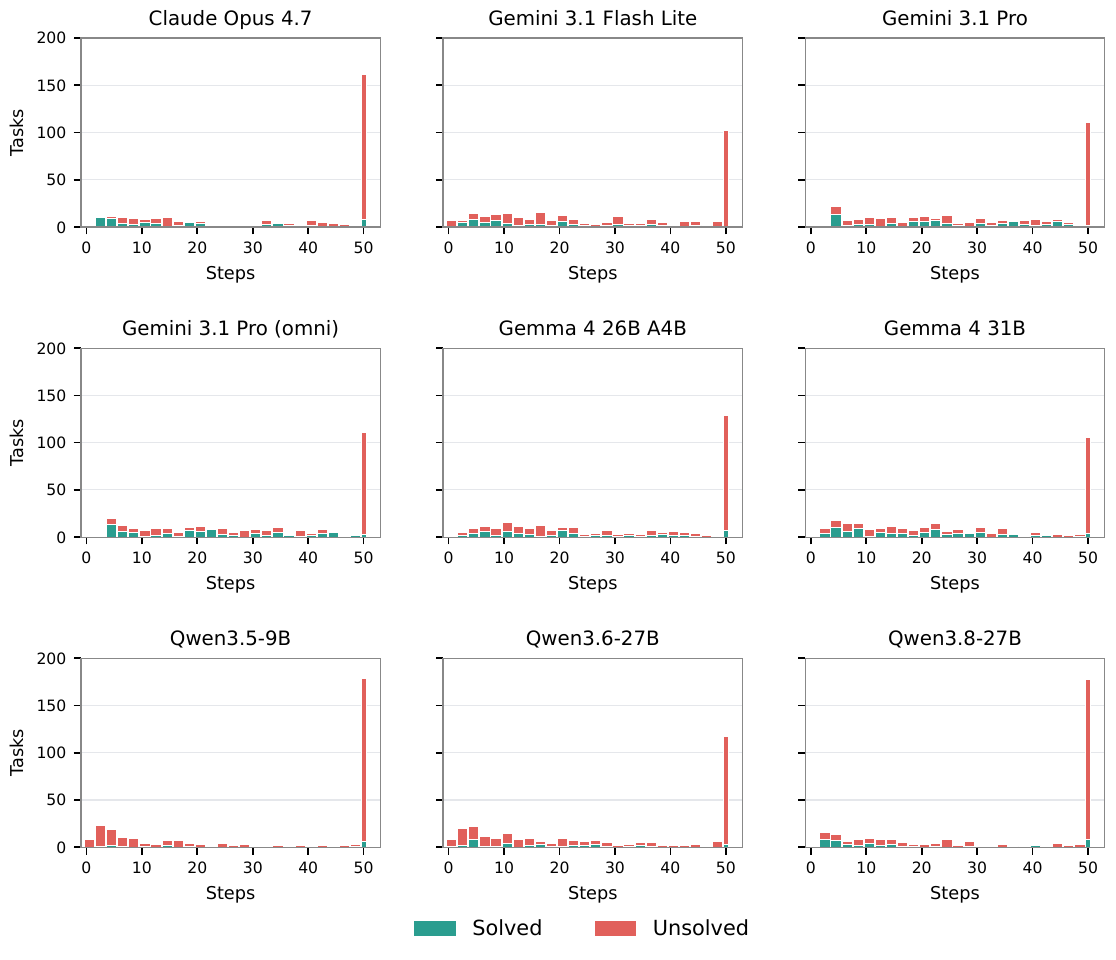}
    \caption{Trajectory length distributions of GenericAgent variants for solved and unsolved tasks on the WebArena-Pro Test split (300 tasks per model), with a 50-step budget. These variants include additional results of Table~\ref{tab:results-additional}.}
    \label{fig:steps_per_model}
\end{figure}

Model-level differences extend beyond trajectory length. 
Table~\ref{tab:model-failure-evidence} provides qualitative examples of models' failures.
Gemini 3.1 Pro has the lowest recorded failure rate. 
Its qualitative failure examples include omitted artifacts and incorrect target entities. 
Both Gemma models exhibit premature or empty finalizations in the reviewed examples. 
Qwen 3.6 27B solves more tasks than Qwen 3.5 9B, but still exhibits missed end-state actions and incorrect entity or value selections. 

Constraint tracking and evidence verification provide further explanations. An exact-constraint error occurs when the agent follows a plausible but task-invalid condition, such as filtering a Metabase category as \texttt{Gadget} instead of the required \texttt{Electronics}, or selecting a flight without satisfying the requested price, date, or currency conditions. Verification failures arise when agents answer from titles, metadata, or nearby content without inspecting the requested video, song, image, or document. Table~\ref{tab:failure-modes-websites} illustrates these overlapping patterns.

\begin{table}[t]
\centering
\small
\caption{Evidence for model-level failure claims in test runs for GenericAgent.}
\label{tab:model-failure-evidence}
\begin{tabular}{@{}p{0.27\linewidth} p{0.69\linewidth}@{}}
\toprule
\textbf{Failure mode} & \textbf{Evidence} \\
\midrule

Gemini Pro omits required artifacts. &
\textbf{Task 225, Peppermint $\rightarrow$ Notion:}
The task required a Notion page summarizing a VPN ticket and naming its owner. All 50 actions remained in Peppermint; the agent never opened Notion or created the page. \\

Gemini Pro selects incorrect target entities. &
\textbf{Task 232, PipeCRM $\rightarrow$ Thunderbird:}
The task required emailing Leon Green at \texttt{leon@example.com}. The agent instead sent the email to \texttt{leon@cedarstone.test}. The recipient rubric failed, although the weekend-question content passed. \\

Both Gemma models exhibit empty finalizations. &
\textbf{Task 229, PipeCRM:}
Gemma 4 26B A4B and Gemma 4 31B both executed \texttt{click('739')} followed by \texttt{done()}. The requested deal's close date, February 21, 2025, and status, \texttt{OPEN}, were visible, but neither model sent an answer.  \\

Qwen 3.6 misses required end-state actions. &
\textbf{Task 22, Acidwave:}
The agent created \texttt{VOCALS\_CONFIRMED} and added ``STAYIN' AT HOME - VOCALS,'' then called \texttt{done()} after seven actions. The required ``FLYING AWAY (VOCALS)'' was absent. \\

Qwen 3.6 selects incorrect entities or values. &
\textbf{Task 52, Firefly $\rightarrow$ Thunderbird:}
The task required emailing \texttt{bob@techcorp.local} that Spotify Premium costs C\$11.99/month. The agent used \texttt{bob@example.com} and \$9.99, clicked Send, and reported completion. \\

\bottomrule
\end{tabular}
\end{table}

Figure~\ref{fig:steps_per_model} shows trajectory lengths for solved and unsolved tasks. The accumulation of unsuccessful runs at 50 steps highlights difficulty completing tasks within the available budget. Across the 2,400 Test runs, 878 of the 1,683 unsuccessful runs (52.2\%) reached the 50-step limit. 
Short trajectories require a different interpretation: they can represent either early failure or successful completion of a short lookup or a task supported by a media tool. Trajectory length alone is therefore insufficient to diagnose premature termination.

\begin{table}[t]
\centering
\caption{Representative failure patterns and illustrative evidence of GenericAgent variants.}
\label{tab:failure-modes-websites}
\begin{tabular}{p{0.23\linewidth} p{0.23\linewidth} p{0.46\linewidth}}
\toprule
\textbf{Failure pattern} & \textbf{Example website(s)} & \textbf{Illustrative evidence} \\
\midrule
Budget exhaustion &
Across applications &
Of Claude Opus 4.7's 224 unsuccessful Test runs, 155 reached the 50-step limit. \\

Incomplete workflow / navigation failure &
Nextcloud, Mattermost &
An agent failed to open the budget file in Nextcloud and never navigated to Mattermost to message Jane. \\

Wrong answer or entity &
Acidwave &
An agent answered \texttt{00:00} for a folk-song duration task whose expected answer was \texttt{33:22}. \\

Filter, sort, or exact-constraint error &
Metabase &
An agent built a product-rating table but filtered \texttt{Gadget} instead of the required category \texttt{Electronics}. \\

Skipped verification &
YT-Lite &
An agent answered using a different Warhammer video without inspecting the requested Warhammer III video. \\

Premature termination &
Peppermint &
An agent declared a database-comment task infeasible without opening the target discussion. \\
\bottomrule
\end{tabular}
\end{table}


\subsection{Coding Agents Failure Modes}

We analyze the five coding-agent configurations on the 300-task WebArena-Pro Test split: Claude Code, Codex, and OpenCode with Gemini 3.1 Pro, Kimi 3.1 Pro, and Qwen 3.8 Max Prime.
Table~\ref{tab:coding-agent-failure-evidence} illustrates representative failure modes.
Across the 1,500 runs, 375 of the 633 graded failures (59.2\%) received partial reward.
Representative failures involve incomplete workflows or missing final state changes: agents prepared records without saving them, created files locally without placing them in the target application, or returned answers instead of updating the requested page.
Stale references, obstructed clicks, and ineffective recovery further disrupted execution.
Other observed failures involve cross-site coordination, visual event counting, and temporal localization despite successful media access.
Short trajectories sometimes reflect clarification requests or explicit refusals rather than premature abandonment.

\begin{figure}[t]
    \centering
    \includegraphics[width=\linewidth]{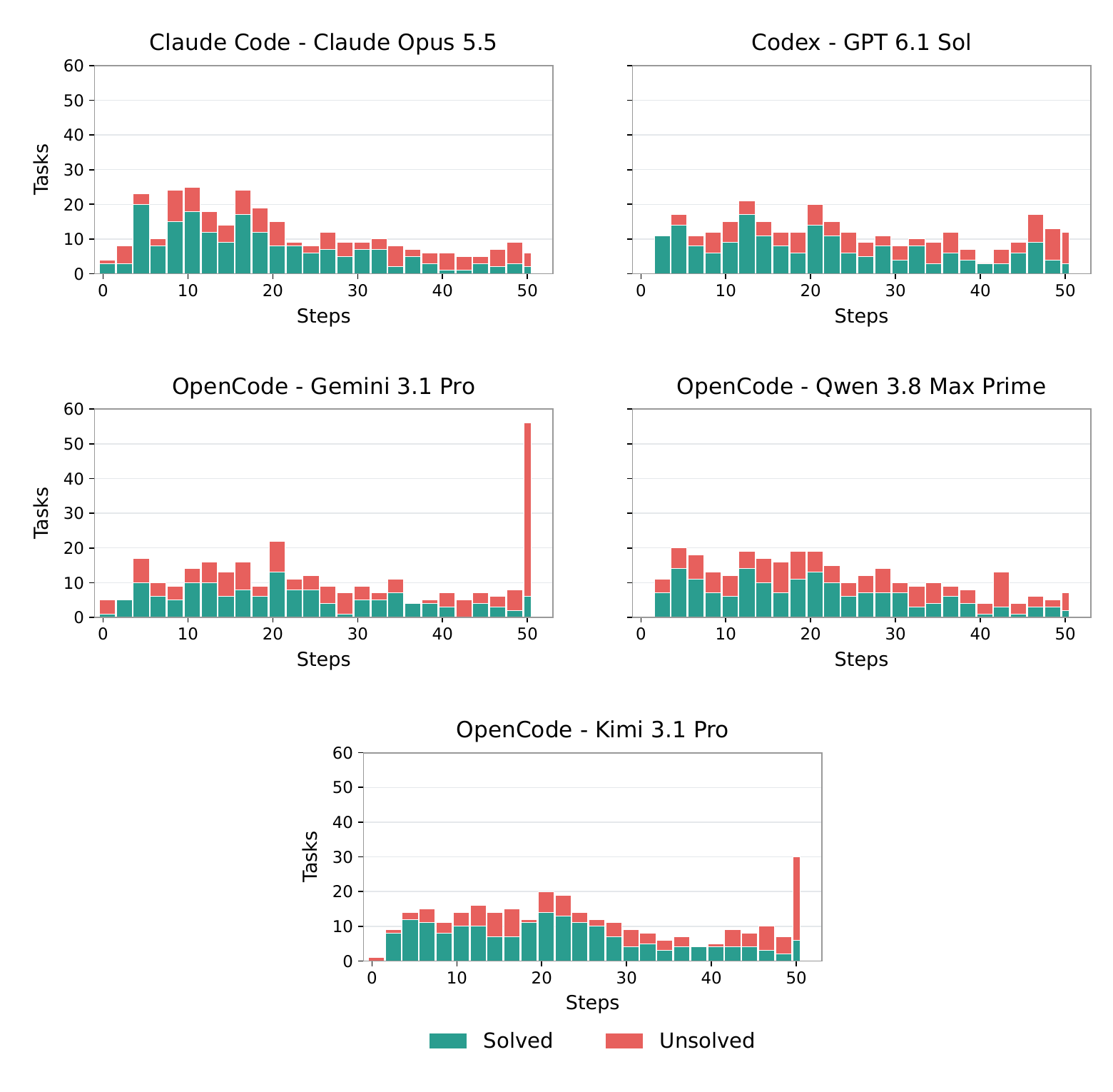}
    \caption{Trajectory length distributions of coding agents for solved and unsolved tasks on the WebArena-Pro Test
split (300 tasks per model), with a 50-step budget.}
    \label{fig:steps_per_model_coding}
\end{figure}

Figure~\ref{fig:steps_per_model_coding} shows trajectory lengths for solved and unsolved tasks across all five configurations.
Of the 633 unsuccessful runs, 92 (14.5\%) reached the 50-step limit.
The proportion was highest for OpenCode with Gemini 3.1 Pro (32.1\%) and Kimi 3.1 Pro (20.3\%), compared with 8.0\% for Codex, 3.8\% for OpenCode with Qwen 3.8 Max Prime, and 3.5\% for Claude Code.
Most unsuccessful trajectories therefore end before exhausting the step budget.

\begin{table}[t]
\centering
\small
\caption{Examples of observed task failures across five coding-agent configurations.}
\label{tab:coding-agent-failure-evidence}
\begin{tabular}{@{}p{0.27\linewidth} p{0.69\linewidth}@{}}
\toprule
\textbf{Failure mode} & \textbf{Evidence} \\
\midrule

Incorrect absence claim. &
\textbf{Claude Code / Opus 5.5, Task 178, Notion:}
Asked to publish the ``my summer vacation'' page, the agent concluded that
the page did not exist and stopped without publishing it. The other
agents successfully published the requested page in their selected runs. \\

Local artifacts not fully delivered to the application. &
\textbf{Codex / GPT 6.1 Sol, Task 147, MyDrive:}
The agent generated all six requested algorithm files locally, but did not
complete their creation in MyDrive. Its final answer acknowledged that
\texttt{Q1\_javascript.js} was still missing from the target folder. \\

Budget exhaustion before committing the result. &
\textbf{OpenCode / Kimi 3.1 Pro, Task 046, ERPNext:}
The agent gathered order and contact information from Metabase and PipeCRM
and prepared a Sales Order. It reached 50 steps before saving; the terminal
screenshot showed the order marked ``Not Saved.'' \\

Incorrect visual event counting. &
\textbf{OpenCode / Gemini 3.1 Pro, Task 317, YT-Lite:}
Asked how many times a rapper appeared holding a Sprite, the agent completed
12 frame-extraction calls but answered \texttt{11}, whereas the rubric
required \texttt{9}. Media access succeeded, but the final count was incorrect
under the grading criterion. \\

\bottomrule
\end{tabular}
\end{table}

\section{Judge Models}
\label{app:judge}

We compare three candidate judges, Qwen 3.6-27B, Gemini-3.1 Pro, and Kimi 2.6, against human annotations. We select 50 tasks and collect one trajectory per task from each of six agent models, for 300 trajectories in total. Human annotators label each rubric of these trajectories as satisfied or not, and we report the rate at which each judge agrees with the human label. Table~\ref{tab:judge_human_agreement} shows the results. All three judges reach similar pooled agreement, between 89.5\% and 90.6\%. We choose Qwen 3.6-27B because it performs comparably to the other judges while being a relatively small open-weight model. It can therefore be hosted locally at a much lower cost, and its fixed weights keep the evaluation reproducible over time, unlike judges served only through a commercial API.

\begin{table*}[t]
\centering
\small
\caption{Rubric-level agreement between each judge and human annotations. Each row corresponds to the agent model that produced the trajectories, with 50 trajectories and about 130 annotated rubrics per agent. The pooled row averages the six agents.}
\label{tab:judge_human_agreement}
\begin{tabular}{lrrr}
\toprule
    & \multicolumn{3}{c}{\textbf{Judge Agreement Rate (\%)}} \\
\cmidrule(lr){2-4}
    \textbf{Agent Model}
    & \textbf{Qwen 3.6 27B}
    & \textbf{Gemini 3.1 Pro}
    & \textbf{Kimi 2.6} \\
\midrule
Gemini 3.1 Pro & 96.9 & 92.3 & 94.6 \\
Gemini 3.1 Flash Lite & 87.7 & 91.5 & 90.8 \\
Claude Opus 4.7 & 93.8 & 92.2 & 92.2 \\
Gemma4 31B & 83.1 & 85.4 & 86.9 \\
Qwen 3.5 9B & 92.3 & 92.3 & 93.1 \\
Qwen 3.6 27B & 83.1 & 89.2 & 86.2 \\
\midrule
Pooled & 89.5 & 90.5 & 90.6 \\
\bottomrule
\end{tabular}
\end{table*}

\section{Websites}
\label{app:websites}

WebArena-Pro includes 20 containerized
website environments. Of these, 9 are built by adapting existing open-source
applications with benchmark-specific seeded data and reset logic, while 11 are
custom websites developed with AI assistance and human review. We provide details about the sources and Docker image sizes in Table \ref{tab:webarena-pro-docker-image-sizes}. Additionally, in WebArena-Pro, there is no data is seeded after the container creation. For most websites, the data is seeded when building the image; the container can then be built fast. Exceptions are made for ERP-Next, which initializes the containers with data when it starts.

\paragraph{Media datasets.}
Some WebArena-Pro environments required media-heavy seed datasets that are stored outside Git and baked into Docker images at build time. For Acidwave, we created a music catalog with CC-licensed MP3 assets and accompanying metadata, including title, artist, album, genre, duration, play counts, license, attribution, and source URL; the current seed references 50 unique audio files. For \texttt{yt-lite}, we used samples from \href{https://huggingface.co/datasets/HuggingFaceFV/finevideo}{FineVideo}: videos were decoded locally, paired with generated thumbnails, and transcoded to 720p HEVC/H.265 for compression before being packaged in the yt-lite image. For Immich, we sampled photos from \href{https://huggingface.co/datasets/bitmind/open-images-v7}{Open Images V7}, assigned them across the admin and agent accounts, manually curated them to remove low-quality, inappropriate, duplicate, or privacy-sensitive images, and then baked the uploaded photos, thumbnails, database state, and ML cache into the final image.

\paragraph{Image Sizes.}
Applications' image sizes in Table~\ref{tab:webarena-pro-docker-image-sizes} are the sums of compressed layer sizes for the current \texttt{linux/amd64} manifests of
\texttt{docker.io/webarenapro/<website>:latest}, queried from Docker Hub and verified against the registry manifests.
Values use decimal units ($1\,\mathrm{MB}=10^6$ bytes; $1\,\mathrm{GB}=10^9$ bytes) and are rounded to three significant figures. They exclude manifest/configuration metadata and do not represent uncompressed installed sizes.

\paragraph{App Licenses.}
Licenses provided in Table~\ref{tab:webarena-pro-docker-image-sizes} refer to original application code or the named upstream software. Dependencies, datasets, media, and other bundled components retain their respective licenses.

\begin{table}[t]
\centering
\caption{Compressed Docker image sizes, upstream sources, and application licences
for WebArena-Pro website environments.}
\label{tab:webarena-pro-docker-image-sizes}
\small
\resizebox{\linewidth}{!}{
\begin{tabular}{llll}
\toprule
Website & Construction basis / upstream repository
& Application/upstream license & Compressed image size \\
\midrule
\texttt{acidwave} & AI-assisted custom
& MIT & 424 MB \\
\texttt{erp-next} & \href{https://github.com/frappe/erpnext}{Open-source ERPNext/Frappe}
& GPL-3.0 (ERPNext); MIT (Frappe) & 627 MB \\
\texttt{firefly} & \href{https://github.com/firefly-iii/firefly-iii}{Open-source Firefly III}
& AGPL-3.0 & 347 MB \\
\texttt{flight-search} & AI-assisted edited from \href{https://github.com/Jonty/airline-route-data}{airline-route-data}
& MIT & 64.0 MB \\
\texttt{immich} & \href{https://github.com/immich-app/immich}{Open-source Immich}
& AGPL-3.0 & 7.03 GB \\
\texttt{jobportal} & AI-assisted edited from \href{https://github.com/aneebnaqvi15/jobportal}{jobportal}
& MIT & 85.6 MB \\
\texttt{mattermost} & \href{https://github.com/mattermost/mattermost}{Open-source Mattermost}
& AGPL-3.0 / Apache-2.0 & 647 MB \\
\texttt{metabase} & \href{https://github.com/metabase/metabase}{Open-source Metabase}
& AGPL-3.0 & 1.05 GB \\
\texttt{minerva-lite} & AI-assisted custom
& MIT & 48.8 MB \\
\texttt{mydrive} & AI-assisted custom
& MIT & 624 MB \\
\texttt{nextcloud} & \href{https://github.com/nextcloud/server}{Open-source Nextcloud}
& AGPL-3.0 & 1.94 GB \\
\texttt{notion} & AI-assisted custom
& MIT & 380 MB \\
\texttt{password-manager} & AI-assisted custom
& MIT & 58.6 MB \\
\texttt{penpot} & \href{https://github.com/penpot/penpot}{Open-source Penpot}
& MPL-2.0 & 625 MB \\
\texttt{peppermint} & \href{https://github.com/Peppermint-Lab/peppermint}{Open-source Peppermint}
& AGPL-3.0 & 335 MB \\
\texttt{pipedcrm} & AI-assisted custom
& MIT & 96.6 MB \\
\texttt{thunderbird} & \href{https://github.com/roundcube/roundcubemail}{Roundcube} + \href{https://github.com/dovecot/core}{Dovecot} webmail stack
& GPL-3.0+; LGPL-2.1 / MIT & 128 MB \\
\texttt{travelmap} & AI-assisted custom
& MIT & 79.4 MB \\
\texttt{x-lite} & AI-assisted custom
& MIT & 139 MB \\
\texttt{yt-lite} & AI-assisted custom
& MIT & 30.1 GB \\

\bottomrule
\end{tabular}
}

\end{table}

\begin{table*}[p]
    \centering
    \small
    \caption{Function-call frequencies across the 300 selected Test
    trials per coding agent. Entries are absolute invocation counts,
    including free observations, rather than budgeted steps.}
    \label{tab:coding-agent-call-frequencies}
    \setlength{\tabcolsep}{5pt}
    \renewcommand{\arraystretch}{1.02}

    \resizebox{\linewidth}{!}{
    \begin{tabular}{@{}lrrrrr@{}}
        \toprule
        \textbf{Function}
        & \shortstack{\textbf{Claude Code}\\Opus 5.5}
        & \shortstack{\textbf{Codex}\\GPT 6.1 Sol}
        & \shortstack{\textbf{OpenCode}\\Gemini 3.1 Pro}
        & \shortstack{\textbf{OpenCode}\\Qwen 3.8 Max Prime}
        & \shortstack{\textbf{OpenCode}\\Kimi 3.1 Pro} \\
        \midrule

        \multicolumn{6}{@{}l}{\textit{Browser tools}} \\
        \texttt{browser\_snapshot}
        & 296 & 1\,036 & 423 & 312 & 300 \\
        \texttt{browser\_take\_screenshot}
        & 76 & 89 & 8 & 30 & 12 \\
        \texttt{browser\_find}
        & 1\,307 & 809 & 357 & 273 & 305 \\
        \texttt{browser\_navigate}
        & 1\,753 & 954 & 1\,099 & 1\,466 & 1\,153 \\
        \texttt{browser\_navigate\_back}
        & 7 & 18 & 53 & 13 & 40 \\
        \texttt{browser\_tabs}
        & 36 & 250 & 248 & 94 & 229 \\
        \texttt{browser\_click}
        & 2\,711 & 4\,196 & 4\,322 & 3\,290 & 3\,534 \\
        \texttt{browser\_hover}
        & 17 & 29 & 52 & 28 & 24 \\
        \texttt{browser\_drag}
        & 9 & 20 & 20 & 6 & 4 \\
        \texttt{browser\_type}
        & 820 & 998 & 1\,399 & 859 & 913 \\
        \texttt{browser\_fill\_form}
        & 196 & 248 & 328 & 259 & 340 \\
        \texttt{browser\_select\_option}
        & 66 & 68 & 64 & 41 & 58 \\
        \texttt{browser\_press\_key}
        & 272 & 328 & 501 & 331 & 347 \\
        \texttt{browser\_handle\_dialog}
        & 6 & 3 & 6 & 6 & 3 \\
        \texttt{browser\_wait\_for}
        & 29 & 8 & 41 & 81 & 339 \\
        \texttt{browser\_file\_upload}
        & 7 & 6 & 6 & 6 & 17 \\
        \texttt{browser\_mouse\_click\_xy}
        & 118 & 59 & 166 & 156 & 140 \\
        \texttt{browser\_mouse\_move\_xy}
        & 32 & 36 & 16 & 25 & 41 \\
        \texttt{browser\_mouse\_drag\_xy}
        & 42 & 44 & 75 & 38 & 55 \\
        \texttt{browser\_mouse\_down}
        & 3 & 3 & 8 & 4 & 10 \\
        \texttt{browser\_mouse\_up}
        & 3 & 4 & 6 & 3 & 9 \\
        \textbf{Browser subtotal}
        & \textbf{7\,806} & \textbf{9\,206}
        & \textbf{9\,198} & \textbf{7\,321} & \textbf{7\,873} \\
        \addlinespace

        \multicolumn{6}{@{}l}{\textit{Media and submission tools}} \\
        \texttt{media\_transcribe}
        & 391 & 387 & 249 & 234 & 427 \\
        \texttt{media\_extract\_frames}
        & 75 & 81 & 57 & 121 & 59 \\
        \texttt{webarena\_submit}
        & 298 & 292 & 285 & 260 & 299 \\
        \textbf{Media/submission subtotal}
        & \textbf{764} & \textbf{760}
        & \textbf{591} & \textbf{615} & \textbf{785} \\
        \addlinespace

        \multicolumn{6}{@{}l}{\textit{Native coding-agent tools}} \\
        Shell (Bash / bash / exec\_command)
        & 341 & 6 & 1\,126 & 428 & 43 \\
        File read (Read / read)
        & 12 & 0 & 17 & 6 & 18 \\
        File write (Write / write)
        & 0 & 0 & 13 & 9 & 18 \\
        File edit (Edit / edit)
        & 0 & 0 & 0 & 1 & 7 \\
        Patch application (apply\_patch)
        & 0 & 3 & 0 & 0 & 0 \\
        File search (glob)
        & 0 & 0 & 0 & 1 & 20 \\
        Text search (grep)
        & 0 & 0 & 0 & 3 & 2 \\
        Tool discovery (ToolSearch)
        & 394 & 0 & 0 & 0 & 0 \\
        Subagent invocation (Agent / Task / task)
        & 2 & 0 & 5 & 10 & 10 \\
        Task planning (todowrite)
        & 0 & 0 & 0 & 13 & 151 \\
        Invalid-call handler (\texttt{invalid})
        & 0 & 0 & 0 & 12 & 0 \\
        \textbf{Native-tool subtotal}
        & \textbf{749} & \textbf{9}
        & \textbf{1\,161} & \textbf{483} & \textbf{269} \\
        \midrule

        \textbf{Total (excluding orchestration)}
        & \textbf{9\,319} & \textbf{9\,975}
        & \textbf{10\,950} & \textbf{8\,419} & \textbf{8\,927} \\
        Mean calls per task
        & 31.1 & 33.3 & 36.5 & 28.1 & 29.8 \\
        \midrule

        \multicolumn{6}{@{}l}{
            \textit{Codex orchestration wrappers (excluded above)}
        } \\
        \texttt{exec}
        & 0 & 7\,199 & 0 & 0 & 0 \\
        \texttt{wait}
        & 0 & 311 & 0 & 0 & 0 \\
        \bottomrule
    \end{tabular}
    }
\end{table*}


\end{document}